\documentclass[]{gtech}
\usepackage{amssymb}
\usepackage{bigdelim}
\usepackage{todonotes}
\usepackage{longtable}
\usepackage{tabularray}
\usepackage{wrapfig}
\usepackage[most]{tcolorbox} 
\usepackage{colortbl}
\usepackage{xcolor}
\usepackage{url}
\usepackage[font=small]{caption} 

\usepackage[utf8]{inputenc} 
\usepackage[T1]{fontenc}    
\usepackage{hyperref}       
\usepackage{url}            
\usepackage{booktabs}       
\usepackage{amsfonts}       
\usepackage{nicefrac}       
\usepackage{microtype}      
\usepackage{lipsum}		
\usepackage{graphicx}
\usepackage{natbib}
\usepackage[frozencache,cachedir=.]{minted}
\usepackage{doi}
\usepackage{color}
\usepackage{arydshln}
\usepackage{amsmath}
\graphicspath{ {./images/} }
\usepackage{multirow}       
\usepackage{listings}       
\usepackage{tcolorbox}      
\usepackage{makecell} 
\usepackage{multicol}

\renewcommand{\title}[1]{\newcommand{\titlelist}{{\Large\fontfamily{optimistic}\selectfont #1}}}

\definecolor{LightGray}{gray}{0.99}
\definecolor{grey1}{RGB}{247,202,173}
\definecolor{grey2}{RGB}{255,241,204}
\definecolor{grey3}{RGB}{225,240,217}
\definecolor{grey4}{RGB}{218,227,243}
\definecolor{grey5}{RGB}{244,247,251}

\newcommand{\prompt}[1]{\rowcolor{grey3!30} #1  \\}
\newcommand{\functionfirst}[1]{\rowcolor{grey2!10} #1  \\}
\newcommand{\functionsecond}[1]{\rowcolor{grey2!10}  #1  \\}
\newcommand{\breadthquestion}[1]{\rowcolor{grey4!70}  #1  \\}
\newcommand{\breadthanswer}[1]{\rowcolor{grey4!30}  #1  \\}
\newcommand{\depthreference}[1]{\rowcolor{grey2!10}  #1  \\}

\title{KAG-Thinker: Interactive Thinking and Deep Reasoning in LLMs via Knowledge-Augmented Generation}

\author[*]{Knowledge Engine Team @Inclusion AI, Ant Group}
\contribution[*]{See Contributions section (Sec.~\ref{sec:contributors}) for full author list.}
\abstract{
In this paper, we introduce KAG-Thinker, which upgrade KAG to a multi-turn interactive thinking and deep reasoning framework powered by a dedicated parameter-light large language model (LLM). Our approach constructs a structured thinking process for solving complex problems, enhancing the  the logical coherence and contextual consistency of the reasoning process in question-answering (Q\&A) tasks on domain-specific knowledge bases (KBs) within LLMs. Following the \textbf{Logical Form} guided retrieval and reasoning technology route of KAG, this framework first decomposes complex questions into independently solvable sub-problems (which are also referred to as logical forms) through \textbf{breadth decomposition}. Each such logical form is represented in two equivalent forms—natural language and logical function—and subsequently classified as either a Knowledge Retrieval or Reasoning Analysis task. Dependencies and parameter passing between these tasks are explicitly modeled via logical function interfaces. In the solving process, the Retrieval function performs retrieval tasks. It retrieves one-hop structured and unstructured information of specified knowledge unit. While the Math and Deduce functions are used to perform reasoning analysis tasks. Secondly, it is worth noting that, in the Knowledge Retrieval sub-problem tasks, LLMs and external knowledge sources are regarded as equivalent KBs. We use the \textbf{knowledge boundary} module to determine the optimal source using self-regulatory mechanisms such as confidence calibration and reflective reasoning, and use the \textbf{depth solving} module to enhance the comprehensiveness of knowledge acquisition.
Finally, instead of utilizing reinforcement learning, we employ supervised fine-tuning with multi-turn interactive thinking and deep reasoning to align the model with our structured reasoning paradigm, thereby avoiding excessive reflection. This is supported by a data evaluation framework and iterative corpus synthesis, which facilitate the generation of detailed reasoning trajectories.
Experimental results show that our model outperforms state-of-the-art trained deep search models on seven benchmark datasets, achieving an average improvement of 4.1\%. 
We further demonstrate its practical effectiveness through a medical Q\&A system integrated with domain-specific knowledge. We train KAG-Med-Thinker model with 14b-parameter based on synthetic medical corpora using the same approach, respectively, and verify their applicability in specialized applications.
}

\gtechdata[Code of KAG-Thinker]{\url{https://github.com/OpenSPG/KAG-Thinker}}
\gtechdata[Code of KAG Framework]{\url{https://github.com/OpenSPG/KAG}}

\begin{document}
\maketitle
\section{Introduction}

Recent advances in large language models (LLMs)~\cite{openai2024gpt4technicalreport,qwen2025qwen25technicalreport} have shown impressive performance in Q\&A tasks. However, their ability to solve multi-hop problems~\cite{cotWei2023}, especially those that require external retrieval of up-to-date information~\cite{DBLP:conf/iclr/JinY0A25}, has certain limitations. Previous work primarily focused on how to use external retrievers, broadly falling into two categories: 
(1) decoupling retrieval from training, and 
(2) autonomously deciding whether to use the retriever during training. 
The first category primarily involves retrieval augmentation generation (RAG) with predefined workflow, the representative works include~\cite{DBLP:conf/nips/LewisPPPKGKLYR020},~\cite{DBLP:conf/acl/TrivediBKS23}, and~\cite{Search-o1}. These models often use the question to retrieve relevant passages, which are then added to the LLM's context. This helps the LLM leverage external knowledge when answering questions. The second category mainly concerns to mimic the deeper and more deliberative reasoning as "slow thinking" ~\cite{DBLP:journals/corr/abs-2505-02665}, a paradigm inspired by human cognitive processes, to determine the timing of retrieval dynamically. Exemplified by~\cite{jin2025searchr1trainingllmsreason},~\cite{sun2025zerosearchincentivizesearchcapability}, and~\cite{wang2025stepsearchignitingllmssearch}. These methods generate a specific token \textcolor{blue}{<search>} that indicates an external search engine query is required. Once the retriever provides results, these are then fed back into the model as input, allowing it to resume content generation. Although these methods improved LLM utilization of retrievers, they do not truly provide a step-by-step approach for solving complex multi-hop problems. Their approaches are more concerned with how LLMs arrive at the correct answer, without ensuring the rigor, stability, or logical consistency of the underlying problem-solving process. Consequently, these methods are often unreliable in high-stakes or sensitive domains such as finance, medicine, and legal applications.

\begin{figure}[htbp]
\centering 
\includegraphics[width=\textwidth]{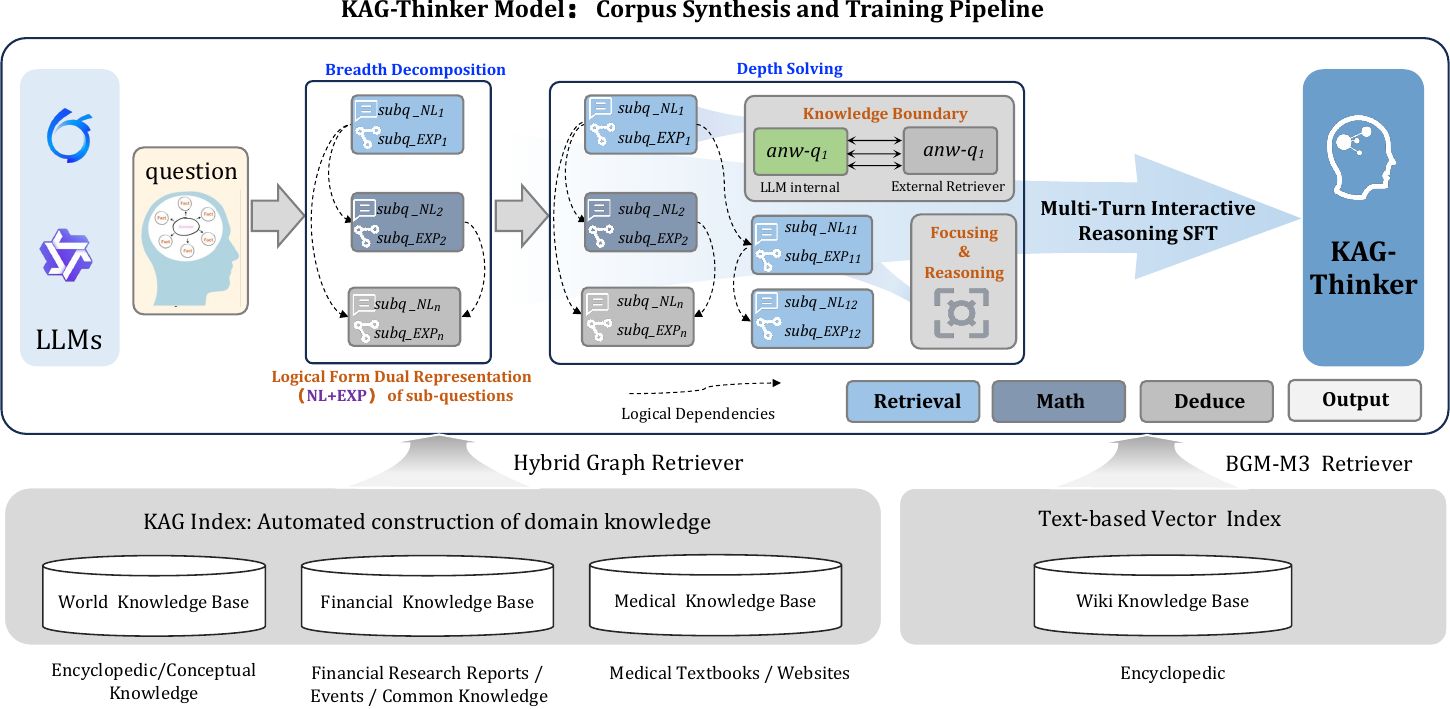} 
\caption{Overview of the KAG-Thinker reasoning model. The corpus synthesis pipeline in the figure from left to right: 1) Decompose the original $question$ into multiple sub-questions with clear logical dependencies and can be solved independently, each sub-question corresponds to a unique logical form, which consists of two components. For example, $logical form_{i}$ consists of $subq\text{\_}NL_{i}$ expressed in natural language and $subq\text{\_}EXP_{i}$ expressed in logical function equivalent. The logical forms are classified into one of $Retrieval$, $Math$, $Deduce$ and $Output$ through function name. 2) Perform knowledge boundary determination and depth solving on $Retrieval$ sub-questions. During depth solving, the focusing and reasoning module reduces noise in retrieved information. 3) Finally, synthesize multi-turn interactive reasoning corpus expressed in natural language and logical function express, and obtain KAG-Thinker Model through supervised fine-tuning of Ling and Qwen's base and instruct models.} 
\label{Fig.KAGThinker_introduction} 
\end{figure}

In order to achieve logical reasoning with more accurate judgments and reduced biases~\citep{Li2025FromS1}, recent work has attempted various approaches, such as 
the atomic reasoning actions decomposition by ~\cite{DBLP:journals/corr/abs-2411-18478} 
, hierarchical and structured reasoning by ~\cite{DBLP:journals/corr/abs-2502-06772} 
, and verify reasoning trajectories with symbolic solver by ~\cite{DBLP:journals/corr/abs-2501-04519} . Moreover, ~\citet{DBLP:journals/tmlr/SumersYN024} propose a cognitive architectures for language agents  with three key dimensions: information storage, action space and decision-making procedure. Benefit from these previous works and inspired by Slow Thinking - which solves complex problem with access to additional computational resources, full attention, and sophisticated logical reasoning ~\cite{DBLP:conf/aaai/BoochFHKLLLMMRS21} - we propose an innovative structured thinking and reasoning model as shown in Figure~\ref{Fig.KAGThinker_introduction}.  It enables deep thinking in LLM by simulating human cognitive mechanisms without reinforcement learning~\citet{deepseekai2025deepseekr1incentivizingreasoningcapability,OpenAIo1}.
We use logical form to decompose complex problems broadly and solve them in depth. Logical form improves the logic, rigor, and stability of the LLM reasoning process. When humans solve complex problems, they only perform external information retrieval on knowledge that they do not know. Similarly to this process, we do not need to perform external knowledge retrieval for all decomposed sub-problems. Therefore, we design an LLM's knowledge boundary determination module to reduce unnecessary external retrieval and mitigate noise from the retrieved content. If the LLM cannot confidently answer a certain sub-question using its own knowledge, we rely on external KBs. However, search results from external KB are often unreliable and noisy. To address this, we created a Focusing-and-Reasoning module to extract the core relevant information. To facilitate a clearer understanding, we will introduce these key innovations in detail in the following paragraphs.

\textbf{Logical Form based Reasoning}. Prior RAG approaches~\cite{DBLP:conf/nips/LewisPPPKGKLYR020,DBLP:conf/acl/TrivediBKS23,Search-o1}, for both questions and sub-questions, are restricted to plain text representations, thereby preventing them from effectively utilizing high-quality structured knowledge. To address this issue, logical form proposes four operation functions tailored to frequently encountered questions (see the Appendix~\ref{appendix:logical_form}). Each operation comprises two components: a natural language segment (Step) and a logical function expression segment (Action). For generic retrievers like E5 and BGE-M3, we can directly employ the content within the logical form's Step. Conversely, for structured knowledge retrievers, the Action portion of the logical form can be utilized. Crucially, logical form facilitates the propagation of dependencies among interdependent sub-problems. For example, consider \textit{"Step1: Who is the director of Hit Parade Of 1947? Action1: Retrieval(s=s1:film[`Hit Parade Of 1947'], p=p1:director, o=o1:director)"} followed by \textit{"Step2: When did \#1 die? Action2: Retrieval(s=o1, p=p2:deathtime, o=o2:deathtime)"}. The variables $\#1$ and $o1$ enable seamless transfer of text and expressions from the outcome of the first planning step. In this manner, the logical form enhances the logic, rigor, and stability of the LLM reasoning.

\textbf{Breadth Decomposition and Depth Solving}.
Direct retrieval often proves insufficient for complex multi-hop questions, failing to provide immediate answers. For example, \textit{"Which film has the director who died first, Hit Parade Of 1947 or Khiladi 420?"}. To address such issues, we have devised a model based on breadth decomposition and depth solving. At a macroscopic level, the decomposition of complex questions into logical forms is generally straightforward and typically does not necessitate recourse to external knowledge. For instance, the aforementioned example can be broken down into five constituent sub-questions: \textit{"Who is the director of Hit Parade Of 1947?", "When did \#1 die?", "Who is the director of Khiladi 420?", "When did \#3 die?"}, and \textit{"Which film was directed by the director who died first according to \#2 and \#4?"}. The resolution of each individual sub-problem, conversely, may entail ranging from $0$ to $M$ (maximum number of searches) retrieval operations, contingent upon its inherent complexity. This in-depth approach to sub-problem solving enables each sub-problem to assimilate adequate external knowledge, thereby reaching a solvable state.

\textbf{Knowledge Boundary Determination}. Most prior approaches~\cite{asai2023selfraglearningretrievegenerate,DBLP:conf/nips/GutierrezS0Y024} initiate retrieval for every question or sub-question, even when the information resides within the LLM's inherent knowledge. However, LLMs do not necessitate external lookups for such straightforward queries. To mitigate this redundancy, we design a knowledge boundary determination module that eliminates unnecessary external retrieval operations. Prior to knowledge boundary determination, the LLM first generates answers to the sub-problems. To ensure the precision of this knowledge boundary determination, we employ two strategies: (1) prompt-based confidence assessment and (2) likelihood-based confidence assessment. Only when both conditions are satisfied is considered that the solution to the sub-problem does not require external retrieval, and the answer generated by the LLM is directly adopted.

\textbf{Focusing and Reasoning}. When the LLM's intrinsic knowledge proves inadequate for resolving a given sub-problem, the necessity for external information retrieval arises. However, there is a large quantity of irrelevant or low-quality external information. Following the acquisition of external data, the model undertakes a rigorous evaluation of its credibility and systematically filters out low-fidelity or irrelevant content. Subsequently, the model synthesizes this curated external information with its internal KB, progressively cultivating a comprehensive understanding of the problem through iterative reasoning cycles and knowledge refinement. This iterative process closely emulates human cognitive mechanisms for knowledge construction, acquisition, and information processing, thus critically underpinning the accuracy and reliability of the final reasoning outcomes.

Ultimately, our research yields a human-like structured iterative thinking and depth solving model offering substantial advancements, notably: (1) Facilitating more effective utilization of high-quality KBs, (2) Contributing to a more logically coherent, rigorous, and stable approach to complex problem solving, (3) Enabling autonomous decisions regarding the reliance on intrinsic model knowledge versus external information, thereby reducing unnecessary queries, and (4) Exhibiting a greater capacity to manage noise within retrieved data.

\section{Related Work}
Retrieval-Augmented Generation (RAG) integrates relevant, real-time information from external KBs through a retrieval mechanism, thereby enhancing its responses with contextual richness and factual grounding~\cite{DBLP:conf/icml/BorgeaudMHCRM0L22,DBLP:conf/cikm/DongLWZXX23,DBLP:conf/acl/LuoETPG0MDSLZL24}. Despite these advantages, a persistent challenge in RAG systems is that retrieved content can often be mixed with irrelevant or erroneous information due to the inherent limitations of current retrieval solutions. Moreover, predefined reasoning adopts structured and modular RAG pipelines that limit flexibility in evolving and open-ended tasks ~\cite{Li2025FromS1}.

To mitigate these challenges, researchers have introduced various components designed to enhance the robustness and effectiveness of RAG systems~\cite{DBLP:journals/corr/abs-2312-10997,DBLP:journals/corr/abs-2404-14851,DBLP:conf/acl/TanD0GFW24}. A key focus in recent studies has been on enabling RAG systems to autonomously determine when to leverage retrieval mechanisms. For example,~\cite{jin2025searchr1trainingllmsreason,sun2025zerosearchincentivizesearchcapability,wang2025stepsearchignitingllmssearch} introduced a special token to explicitly trigger an external search, with the retrieval results subsequently integrated into the model to support content generation. However,~\cite{DBLP:conf/coling/RenWQZ00W025} identified limitations in LLM regarding their ability to accurately distinguish between internal and external knowledge, as well as their difficulty in reconciling conflicts between the two. To address this, they proposed Prior judgment and Poster judgments to evaluate confidence levels and assess response correctness.
In parallel, another significant line of research has focused on developing adaptive strategies to optimize how retrieval is utilized, based on query complexity. ~\cite{DBLP:conf/naacl/JeongBCHP24} developed a classifier to intelligently decide between iterative, single, or no retrieval strategies depending on the complexity of the query. Building upon this idea,~\cite{DBLP:conf/emnlp/IslamRHHJP24} proposed a hybrid adaptive retrieval method: The model initially generates an answer without retrieval and then dynamically decides the need for retrieval based on its confidence, with the aim of balancing performance gains with inference speed. Furthermore,~\cite{DBLP:conf/coling/TangGLDLX25} extended this work by applying a reinforcement learning-based framework to select the most appropriate retrieval strategy for a given query.
While these diverse approaches have significantly advanced the integration and adaptive use of retrieval mechanisms within RAG systems, they currently lack a comprehensive, step-by-step framework for effectively addressing complex multi-hop reasoning problems. This critical area remains an open challenge for further research.

\section{Approach}
\begin{figure}[ht]
\centering 
\includegraphics[width=\textwidth]{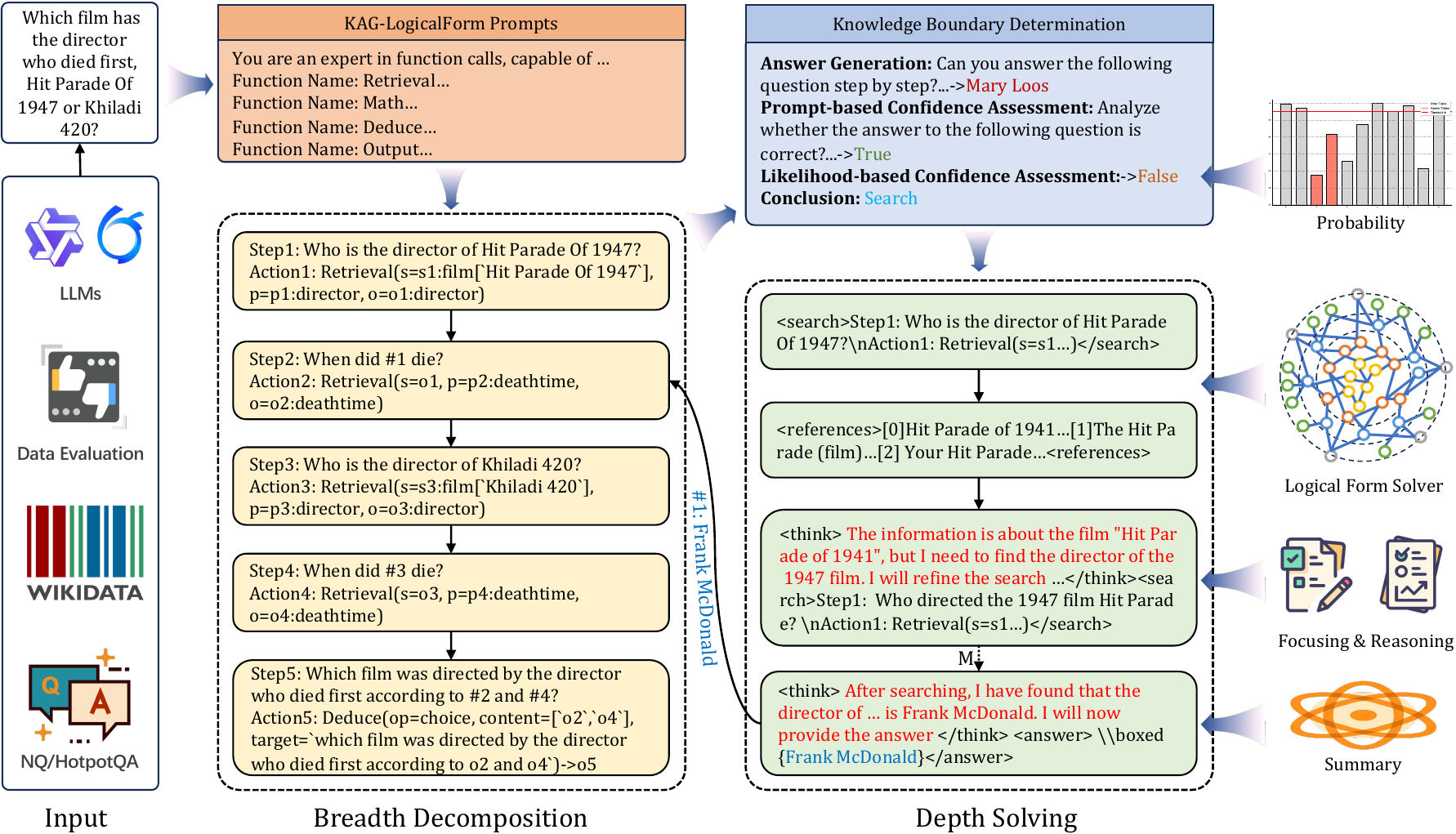} 
\caption{Overview of the problem-solving model that uses human-like reasoning process. During problem breadth decomposition, all sub-problems are obtained in a single decomposition pass, where each sub-problem is an atomic problem that can be solved independently, also known as a logical form. Herein, the terms \textbf{Step} and \textbf{Action} maintain semantic consistency, both denoting such a sub-problem. Within problem breadth decomposition, Step employs $\#n$ for answer propagation of the $n$-th sub-problem, while Action binds variables in logical function (e.g., $o_n$, $s_n$) for variable transmission. By determining the knowledge boundary of the \textit{Retrieval} sub-problem, it is decided whether to utilize the base model's answer or to generate a deep retrieval. During depth solving of sub-problems, the system sequentially executes retrieval, focusing, and reasoning in iterative processes until either the sub-problem answer is obtained or the maximum solving attempt threshold is activated.} 
\label{Fig.KAGThinker_framework} 
\end{figure}
We propose an innovative complex problem solving model, as shown in Figure~\ref{Fig.KAGThinker_framework}. At its core, this model uses logical form to enable both broad problem decomposition and in-depth problem solving. By integrating a knowledge boundary determination module, we minimize unnecessary external queries, balancing the LLM's internal knowledge with external information. For sub-problems that require external information, a dedicated focusing and reasoning module is employed to reduce noise in the retrieved content and extract essential information.

\subsection{Breadth Decomposition and Depth Solving}
Complex multi-hop problems usually need to be broken down to be solved. The key question is how to break them down effectively to best use our existing KBs. We have created logical forms to represent sub-problems. Our method decomposes problems into two parts: broad decomposition, which ensures the main problem and sub-problems stay logical and precise; and in-depth solving, which makes sure \textit{Retrieval} sub-problem gets enough knowledge to be solved.

\subsubsection{Breadth Decomposition}
We decompose complex problems breadth-wise into $n$ atomic granularity sub-problems, with our decomposition template detailed in Table~\ref{tab:breadth_decomposition}. We define four logical form functions (Retrieval, Math, Deduce, and Output), each dedicated to handling specific tasks: retrieval-focused problems, mathematical computation and causal reasoning tasks, and results aggregation functions. Detailed specifications for these modules can be found in Appendix~\ref{appendix:logical_form}. As shown in Table~\ref{tab:breadth_decomposition}, our breadth decomposition divides the original question into five logical forms of atomic granularity, each independently solvable. The dependency variables are propagated between logical forms by function variable $\#n$, $o_n$, and $s_n$. This approach ensures logically coherent question-solving while maintaining consistent sub-question granularity. To ensure compatibility with both natural language and Logical Form retrievers during sub-question solving, each sub-question adopts dual representations, \textbf{Step} and \textbf{Action}, that maintain semantic equivalence.
\renewcommand\arraystretch{1.0}
\begin{table}[ht]
\centering
\small
\setlength\aboverulesep{0pt}\setlength\belowrulesep{0pt}
\begin{tabular}{p{0.97\textwidth}}
\toprule
\prompt{You are an expert in function calls, capable of accurately understanding function definitions and precisely decompose user queries to select appropriate functions to solve problems. The functions are as follows:} 
\functionfirst{Function Name: Retrieval}
\functionsecond{\quad Description: Search for SPO information. S stands for subject, O stands for object, represented as variable\_name...}
\functionsecond{\quad Function Usage: Retrieval(s=s\_alias:type[`name'], p=p\_alias:edge, o=o\_alias:type[name], p.prop=`value', s.prop=`value', o.prop=`value')}
\functionfirst{Function Name: Math}
\functionsecond{\quad Description: Perform calculations, which include set operations such as numerical calculations or sorting...}
\functionsecond{\quad Function Usage: Math(content=[`known conditions' or `o\_alias/s\_alias'], target=`goal to be solved')->math\_alias }
\functionfirst{Function Name: Deduce}
\functionsecond{\quad Description: Inference refers to the process of inferring search or calculation results to answer questions...}
\functionsecond{\quad Function Usage: Deduce(op=judgement|entailment|extract|choice|multiChoice, content=[`known conditions' or `o\_alias/s\_alias'], target=`goal to be solved')->deduce\_alias}
\functionfirst{Function Name: Output}
\functionsecond{\quad Description: Directly output A, B, ... as the answer, where A and B are variable names referencing previous...}
\functionsecond{\quad Function Usage: Output(A,B,...)}
\prompt{Please, based on the definition of the above function, decompose the user question into one or multiple logical steps, outputting the execution plan for each step along with the corresponding action. Please note:
Step: Accurately point out the logical thinking process of the question, and use \#1 to refer to the solution result of Step1, \#2 to refer to the solution result of Step2, and so on
Action: Indicate exactly the function you selected and its parameters.} \hline
\breadthquestion{\textbf{Question:} Which film has the director who died first, Hit Parade Of 1947 or Khiladi 420?} \hline
\breadthanswer{\textbf{Output:} \textcolor{red}{<think>}This is a comparison problem. To address the question of which film, between "Hit Parade Of 1947" and "Khiladi 420" was directed by the director who passed away first, we will undertake a chronological investigation. Our initial step involves identifying the director of "Hit Parade Of 1947". Following this, we will ascertain the director of "Khiladi 420". With both directors' identities established, we will then determine which of these individuals died earlier. This chronological analysis will enable us to connect the director's lifespan to the films they directed, ultimately revealing which film was associated with the director who died first. By meticulously comparing the timelines of the directors' lives, we ensure that our conclusion is based on accurate historical data.\textcolor{red}{</think>}\texttt{$\backslash$n}\texttt{$\backslash$n}\textcolor{teal}{<answer>}Step1: Who is the director of Hit Parade Of 1947?\texttt{$\backslash$n}Action1: Retrieval(s=s1:film[`Hit Parade Of 1947'], p=p1:director, o=o1:director)\texttt{$\backslash$n}Step2: When did \#1 die?\texttt{$\backslash$n}Action2: Retrieval(s=o1, p=p2:deathtime, o=o2:deathtime)\texttt{$\backslash$n}Step3: Who is the director of Khiladi 420?\texttt{$\backslash$n}Action3: Retrieval(s=s3:film[`Khiladi 420'], p=p3:director, o=o3:director)\texttt{$\backslash$n}Step4: When did \#3 die?\texttt{$\backslash$n}Action4: Retrieval(s=o3, p=p4:deathtime, o=o4:deathtime)\texttt{$\backslash$n}Step5: Which film was directed by the director who died first according to \#2 and \#4?\texttt{$\backslash$n}Action5: Deduce(op=choice, content=[`o2',`o4'], target=`which film was directed by the director who died first according to o2 and o4')->o5 \textcolor{teal}{</answer>}}
\bottomrule
\end{tabular}
\caption{Instruction template for breadth decomposition. The template initiates with problem positioning and analysis to determine decomposition steps, ultimately generating all atomic sub-problems through single-pass decomposition. Sub-question dependencies are propagated via $\#n$ in natural language Steps and referenced through logical function variables (e.g., $o_n$, $s_n$) in Actions.}
\label{tab:breadth_decomposition}
\end{table}
\subsubsection{Depth Solving}
Constrained by sub-problem representations and retriever capabilities, many sub-problems defy resolution through single-turn retrievals. We have engineered a depth-oriented resolution strategy for \textit{Retrieval} sub-problem, as detailed in the right part of Figure~\ref{Fig.KAGThinker_framework}. During depth solving, we prompt the LLM to iteratively search across multi-level and multi-dimensional contexts. The iterative process continues until one of the following conditions is met: (1) the maximum number of turns is reached, or (2) the model generates a conclusive response enclosed between the designated answer tokens \textcolor{teal}{<answer>} and \textcolor{teal}{</answer>}. Before conducting depth solving of the \textit{Retrieval} sub-problem, we first perform a knowledge boundary determination on the sub-problem to assess whether the LLM can directly answer it. If the LLM can confidently answer the current sub-problem, there is no need for depth solving, and the answer can be directly obtained; otherwise, an depth solving process will be initiated. After retrieving relevant content, we conduct focusing and reasoning analysis, allowing the model to determine contextually whether to initiate the next action. If a subsequent action is required, a new Logical Form is generated; otherwise, the sub-question answer is produced. This process iterates interactively until either the maximum number of turns is reached or the sub-question answer is obtained.

\subsection{Knowledge Boundary Determination}
When addressing complex problems, LLMs commonly decompose them into a series of constituent sub-problems. For the resolution of these sub-problems, two principal strategies are conventionally employed: (1) direct leveraging of the model's intrinsic KB, and (2) external information retrieval to augment the solution's correctness and scope. However, each of these approaches presents distinct inherent limitations:

\textbf{\textit{Limitations of Internal Knowledge Utilization.}}
The training paradigm based on maximum likelihood estimation during the pre-training phase induces a strong generative propensity within the model's parameter space. Consequently, when confronted with ambiguous or unknown information, the model can generate erroneous or fabricated responses. This phenomenon, widely recognized as the hallucination problem, stems from the model's unwarranted epistemic overconfidence and significantly compromises the veracity of the final output.

\textbf{\textit{Limitations of External Knowledge Retrieval.}}
External knowledge retrieval is indeed capable of mitigating limitations stemming from a model's intrinsic KB. However, indiscriminate retrieval across all sub-problems engenders significant inefficiencies, manifesting as superfluous computational expenditure and redundant acquisition of information already encoded within the model's parameters. 

Therefore, our work is mainly focused on reasonably determining the knowledge boundary of the model. To achieve this, we propose a \textbf{Generate First, Then Assess} strategy. Specifically, the model first attempts to directly generate answer to the sub-problem. The prompt is shown in Table \ref{table:prompt_answer}.  

\begin{table}[h]
    \centering 
    \small
    \begin{tabular}{p{13cm}}
        \hline
        \prompt{Can you answer the following question step by step?}
        \prompt{If you can, wrap your answer with <answer>\textbackslash\textbackslash boxed\{your answer\}</answer>.}
        \prompt{If you can't, just reply that based on my internal knowledge, I can't answer this question, I need to retrieve external knowledge.}
        \breadthquestion{\textbf{Question}: Who is the author of The Pequod?}
        \hline
    \end{tabular}
    \caption{Instruction template for solving sub-problem.}\label{table:prompt_answer}
\end{table}

Next, we combine two confidence assessment methods: prompt-based confidence assessment and likelihood-based confidence assessment, to assess the reliability of the generated answer. The advantage of this approach is that it allows the model to first try to use the internal KB to answer the question. For a sub-problem with high enough confidence in the generated answer, this generated answer will be directly used as the final answer. Otherwise, the final answer will be obtained with the help of external knowledge.

\textbf{1) Prompt-based Confidence Assessment. } 
This confidence assessment method involves gauging confidence by asking the model sub-problems and assessing the authenticity of its generated answer. Specifically, after the model answers a sub-problem, the sub-problem and the generated answer are fed back into the model, prompting it to determine whether they match. Through this prompt-based interactive method, the model can leverage its internal knowledge to reflect and verify its generated answer step by step, ultimately generating a judgment of \textit{True} or \textit{False}. The details of the prompt are shown in Table \ref{table:prompt_assess}.

\begin{table}[h]
    \centering 
    \small
    \begin{tabular}{p{13cm}}
        \hline
        \prompt{Analyze whether the answer to the following question is correct?}
        \prompt{Please think step-by-step before arriving at your conclusion.}
        \prompt{If yes, answer True; if no, answer False.}
        \prompt{Wrap your answer with <answer>\textbackslash\textbackslash boxed\{True/False\}</answer>.}
        \breadthquestion{\textbf{Question:} Who is the author of The Pequod?}
        \breadthanswer{\textbf{Answer:} Herman Melville}
        \hline
    \end{tabular}
    \caption{Instruction template for confidence assessment. Where Question represents the sub-problem to be solved, and Answer represents the LLM-generated answer.}\label{table:prompt_assess}
\end{table}

\begin{figure}[h]
    \centering
    \includegraphics[width=\linewidth]{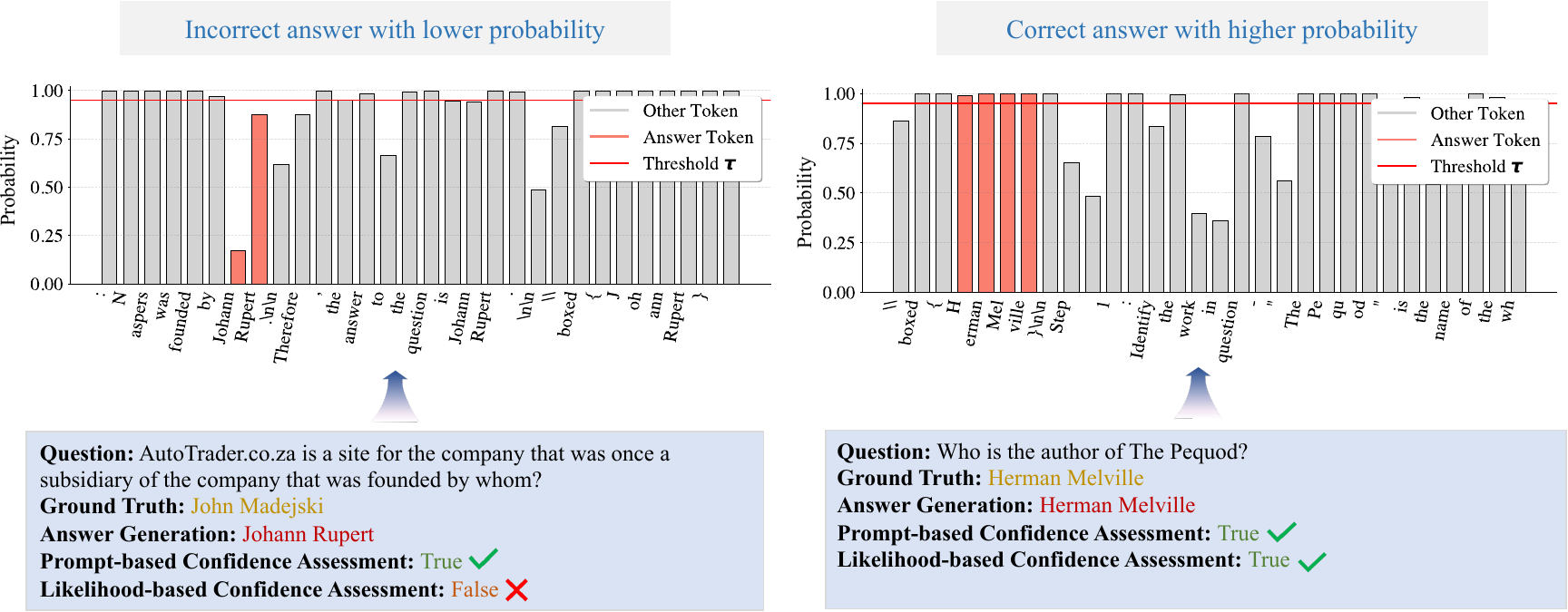}
    \caption{The probability distribution of the answer tokens. The left shows the low confidence answers, and the right shows the high confidence answers. }
    \label{fig:token_probs}
\end{figure}

\begin{table}[b!]
\centering
\small
\setlength\aboverulesep{0pt}\setlength\belowrulesep{0pt}
\begin{tabular}{p{0.97\textwidth}}
\toprule
\prompt{According to the question and references, analyze the relationship between each reference and the question, then summarize whether these references can answer the above questions and give reasons. Wrap the answer and the corresponding reasons in \textcolor{teal}{<answer>}Yes/No\textcolor{teal}{</answer>} and \textcolor{orange}{<reason>}your reason\textcolor{orange}{</reason>}. }\hline
\depthreference{\textbf{References:}}
\depthreference{0. "Splash (U.S. TV series)" Splash (U.S. TV series) Splash is an American competition-based reality show with celebrity diving competitions which broadcast on ABC from March 19 to May 7, 2013. It was based on the Dutch reality franchise ""Celebrity Splash!"" created by Eyeworks for the series ""Stars Jumping on Saturday"" that premiered in 2012. The show was hosted by actor Joey Lawrence and sportscaster Charissa Thompson with former Olympic divers Steve Foley from Australia and David Boudia from the U.S. as judges. Greg Louganis, also a former U.S. Olympic diver, was the diving instructor and mentor to the celebrities. The series was first}
\depthreference{1. "Celebrity Splash!" Celebrity Splash! Celebrity Splash! is a reality television franchise created by Dutch company Eyeworks, started from their Dutch reality show ""Sterren Springen Op Zaterdag"" which premièred in 2012. The franchise involves celebrities diving into the pool. ""Splash"" has its origin and idea from German Olympic-themed variety TV show """" (""TV total Diving""), it was first aired on 16 December 2004, in the ""TV total"" show, on ProSieben and was founded by Stefan Raab and hosted by Sonya Kraus. Other hosts/reporters include Ingolf Lück (2004), Kai Pflaume (2005), Oliver Welke and (2007, 09), Steven Gätjen (2011–12) and (2011–12 reporter). It has}
\depthreference{2. "Celebrity Splash! (Argentinian TV series)" Celebrity Splash! (Argentinian TV series) Splash! is a reality television show which teaches celebrities the art of diving. The first series is featured to start on 11 June 2013, and it will be broadcast by Telefe. Marley Wiebe will be hosting the show, with Pampita Ardohain, Maximiliano Guerra, Miguel Ángel Rodríguez and Mariana Montes as judges. ""Splash"" has its origin and idea from German TV Olympic-themed variety show (""TV Total Diving""), it was first aired on 16 December 2004, in the TV total slot, on ProSieben and was founded by Stefan Raab and hosted by Sonya Kraus. Other hosts/reporters} \hline
\breadthquestion{\textbf{Question:} Who is the host of Splash!} \hline
\breadthanswer{\textbf{Output:}} 
\breadthanswer{\textcolor{teal}{</answer>}Yes\textcolor{teal}{</answer>}}
\breadthanswer{\textcolor{orange}{<reason>}The host of the U.S. TV series Splash is actor Joey Lawrence, alongside sportscaster Charissa Thompson. This is explicitly confirmed in reference 0, which details the U.S. version of the show. References 1 and 2 discuss the Dutch and Argentinian iterations of the franchise, which have different hosts (e.g., Sonya Kraus, Marley Wiebe) and are unrelated to the U.S. version.\textcolor{orange}{</reason>}} 
\bottomrule
\end{tabular}
\caption{Focusing and Reasoning Data Construction Example: By analyzing and evaluating whether the reference materials can sufficiently address the sub-question, the system determines whether to proceed to the next Action or directly generate the final answer.}
\label{tab:focusing_and_reasoning}
\end{table}

\textbf{2) Likelihood-based Confidence Assessment. }
Confidence assessment based on likelihood is a method that evaluates the reliability of generated answers by utilizing the probabilities of output tokens. As shown in Figure \ref{fig:token_probs}, when the model processes a given sub-question input $\boldsymbol{x}$, it first extracts the content within \textbackslash boxed\{\}. Then, this extracted content undergoes relocation, and the relocated result is recorded as $\boldsymbol{y} = \{y_1, y_2, \dots, y_T\}$, where $T$ is the length of the recorded answer sequence. During the generation process, each token $y_t$ ($t = 1, 2, \dots, T$) is associated with a generation probability $P(y_t \mid \boldsymbol{x}, y_{<t})$, which represents the conditional probability of generating $y_t$ given the input and the context of previously generated tokens. The confidence of the generated answer can be evaluated based on these generation probabilities. However, directly taking the average of all token probabilities is not suitable for confidence evaluation, as certain high-frequency tokens (e.g., "the" or "of") tend to have inherently high probabilities during generation. These tokens contribute little to evaluating the model's reliability in generating critical content. To better capture areas of higher uncertainty during generation, we select the lowest generation probability among all tokens and define it as the confidence score $C$ of the result:
\begin{center}
$C = \min \{p(y_{1} \mid \boldsymbol{x}), p(y_{2} \mid \boldsymbol{y}_{<2}, \boldsymbol{x}), \dots, p(y_{T} \mid \boldsymbol{y}_{<T}, \boldsymbol{x})\}$
\end{center}

Once the confidence score $C$ is obtained, we can introduce a predefined threshold $\tau$ to further determine whether the generated answer is considered reliable. If the confidence score $C$ satisfies: $C \geq \tau$, then the confidence of the generated answer is considered \textit{True}; otherwise, it is considered \textit{False}.
\renewcommand\arraystretch{1.0}

After obtaining the confidence results from the two assessment methods, we adopt a dual validation strategy to combine them. Specifically, the final confidence is judged as \textit{True} only when both confidence assessment methods independently yield \textit{True}. This indicates that the model's generated answer is reliable and can be directly utilized. Conversely, if either of the two assessment methods results in \textit{False}, the final confidence is deemed \textit{False}. In this case, the model's generated answer is considered potentially unreliable and requires external knowledge to obtain a more accurate answer.

\subsection{Focusing and Reasoning}
\label{sec:focusing_reasoning}
During knowledge boundary determination, we ascertain whether a sub-question requires retrieval. After executing retrieval, we need to assess if the current retrieval results can adequately address the sub-problem. Therefore, we have designed Focusing and Reasoning modules to fulfill this purpose. Focusing and Reasoning are primarily designed to analyze reference materials, with their data construction methodology as detailed in Table~\ref{tab:focusing_and_reasoning}. Subsequently, based on the answer and reason from the Output section of Table~\ref{tab:focusing_and_reasoning}, we employ LLMs to perform conditional restructuring, ultimately embedding the restructured content into the designated \textcolor{blue}{<think>} within the depth-solving module.

\subsection{Logical Form Solver}
\label{sec.logical_form_solver}
Logical Form, through its dual representation, not only ensures more rigorous logical propagation between sub-problems but also accommodates both natural language retrievers and logical form solvers. The retrieval and reasoning of sub-questions are implemented by Logical Form Solver in KAG framework, which consists of four types of logical form executors: \textbf{Retrieval Executor} is used to retrieve knowledge from an LLM's internal KB or external KB. \textbf{Math Executor} and \textbf{Deduce Executor} are utilized for Reasoning Analysis. Math is used for math or code calculations, and Deduce is an LLM causal reasoning task. \textbf{Output Executor} is used to determine and output the answer to the question. These four executors are used to connect LLMs with external executors, transforming natural language sub-problems into a structured problem-solving process. The grammatical structure of the logical form is explained in Appendix \ref{appendix:logical_form}. Table \ref{tab:solver example} shows an example of using these four executors to solve the problem of \textit{calculating the amount of credit card fraud crime}. During the inference phase, KAG-Thinker generates equivalent representations of natural language and logical function for each logical form (such as Step1 and Action1 in table \ref{tab:solver example}), and then calls the corresponding Executor based on the dependencies.

\begin{table}[b!]
    \small
    \centering
    \setlength\aboverulesep{0pt}\setlength\belowrulesep{0pt}
    \begin{tabular}[l]{p{4.4 cm}|p{6.6 cm}|p{4.2 cm}}
        \toprule
         \multicolumn{3}{p{15.6 cm}}{ {\textbf{Question:} Zhang found a wallet on the street, which contained 5,000 yuan in cash and a credit card. Zhang took the cash and used  the credit card to make purchases at four stores: A, B, C, and D, for 210.4 yuan, 569.2 yuan, 1035.2 yuan, 2044.5 yuan, and 1035 yuan, respectiely. After the incident, the public security authorities determined that Zhang was suspected of embezzlement and credit card fraud. What is the total amount involved in the credit card fraud charge against Zhang?  } } \\
         
       \midrule
       \textbf{Steps} & \textbf{Actions}  & \textbf{Result} \\
       \cline{1-3} 
       \midrule
       1. What's the elements of the offense credit card fraud? 
       & Retrieval(s=s1:offense[credit card fraud], p=p1: found, o=o1:elements) & o1 = “The elements of the offense of credit card fraud include the following points: 1….” \\
       \midrule
       2. Which amounts are considered part of the credit card fraud amount?
       & Deduce(op=extract, content=[o1, Zhang found a wallet on the street…], target=Which amounts are considered part of the credit card fraud amount?)->o2
  &  o2 = “The total amount involved in Zhang’s credit card fraud is 210. 4 yuan, 569.2 yuan, 1035.3 yuan, 2044.5 yuan, and 1035 yuan.”  \\
       \midrule
       3. What's the total amount involved in Zhang’s credit card fraud? & Math(content=[o1, o2], target=What's the total amount involved in Zhang’s credit card fraud.)->o3 &   o3 = “the total amount involved in Zhang’s credit card fraud is 4989.40 yuan.” \\
       \midrule
      
       4. Output \#3.  & Output(o3)
   & Amount: 4894.40 yuan. \\
       \bottomrule
    \end{tabular}
    \caption{An example of the working principle of the Logical Form Solver in the integrated application of four executors:
    It search and references relevant legal provisions by \textbf{Retrieval}, extracts the amount of each illegal use of the credit card in the question by \textbf{Deduce}, calculates the amount involved in the credit card fraud case by \textbf{Math} and finally \textbf{Output} the answer.}
    \label{tab:solver example}
\end{table}

\textbf{Retrieval.} The logical function of Retrieval is defined as $Retrieval$(\textit{$s$, $p$, $o$, $p$.contraints, $p$.constraints, $o$.constraints}). The function variables are composed in the form of Subject-Predicate-Object (SPO) ($s$, $p$, $o$) together with conditional constraints. For example, \textit{"Who was the director of the 2002 film Men in Black?"}, which can be transformed into $Retrieval$ (\textit{$s$: Movie[Men in Black], $p$: DirectedBy, $o$: Person, $s$.ReleaseYear=2002}). The constraints can be applied to $s$, $p$, or $o$. The final retrieval results include the subgraph $SG(s, p, o)$ and chunks related to target variable object $o$.  

In the recently released KAG 0.8, we have enhanced the Hybrid Graph Retriever (HGR)'s knowledge base index management capabilities, with built-in support for basic index types such as KnowledgeUnit, AtomicQuery, Outline, Summary, Chunk, Table, etc. to better support the knowledge structuring of unstructured documents. For detailed introduction, please refer to the Release Note\footnote{KAG v0.8 release note: \url{https://openspg.github.io/v2/blog/recent_posts/release_notes/0.8}}.

\textbf{Deduce.} The logical function of Deduce is defined as $Deduce$ (\texttt{op}, \texttt{content}, \texttt{target}) $\rightarrow y_{d}$, where \texttt{op} specifies the reasoning method, including \textit{extract}, \textit{judgement}, \textit{entailment}, and \textit{choice}. The \texttt{content} field specifies the reasoning context, which includes the information related to the current sub-problem extracted from the user question and the calculation result of the previous logical forms passed through the intermediate variable. The \texttt{target} represents the current objective of the Deduce operation, and $y_{d}$ represents the result variable of the Deduce reasoning.

\textbf{Math.} The logical function of Math is defined as $Math$ (\texttt{content}, \texttt{target}) $\rightarrow y_{m}$, where \texttt{content} and \texttt{target} are similar to Deduce, specifying the known conditions required for the calculation and the computation task, respectively. $y_{m}$ represents the result of the computation.

\textbf{Output.} When the question can be answered using either the model's internal knowledge or the output of previous executors, the output executor will then determine the format and deliver the final answer.

\section{Experiments}
\subsection{Experimental Settings}
\textbf{Benchmarks.} Our experiments are conducted on 7 widely-used datasets. The dataset comprises two primary categories: (1) Single-Hop QA: NQ~\cite{kwiatkowski-etal-2019-natural}, TriviaQA~\cite{joshi-etal-2017-triviaqa}, and PopQA~\cite{mallen-etal-2023-trust}. (2) Multi-Hop QA: HotpotQA~\cite{yang-etal-2018-hotpotqa}, 2WikiMultiHopQA~\cite{ho-etal-2020-constructing}, Musique~\cite{trivedi-etal-2022-musique}, and Bamboogle~\cite{press2023measuringnarrowingcompositionalitygap}. These datasets encompass diverse search and reasoning challenges, thereby enabling comprehensive evaluation of our model. Our evaluation set maintains methodological consistency with established prior works~\cite{chen2025researchlearningreasonsearch,jin2025searchr1trainingllmsreason}.

\textbf{Evaluation Metric.} While EM~\cite{yu2024rankragunifyingcontextranking} provides strict verification of answer precision; however,its binary nature is overly restrictive for open-ended scenarios where semantically equivalent answers may exist legal phrasing differences. The F1 metric therefore serves as a nuanced supplement, particularly valuable when evaluating free-form textual output.

\textbf{Comparison Methods.}
To systematically assess KAG-Thinker's effectiveness, we establish a three-tiered evaluation model comprising the following state-of-the-art baselines:
(1) \textbf{Non-Retrieval Paradigms} Naive Generation: Direct answer synthesis without external knowledge integration; Chain-of-Thought (CoT): Explicit cognitive pathway formalization through sequential reasoning traces~\cite{cotWei2023}. (2) \textbf{Retrieval-Augmented Architectures} Naive RAG: Standard retrieval-generation pipeline without iterative optimization~\cite{DBLP:conf/nips/LewisPPPKGKLYR020}; IRCoT: Multi-cycle retrieval-reasoning coordination with dynamic feedback mechanisms~\cite{DBLP:conf/acl/TrivediBKS23}; Search-o1: Agent-mediated search workflow integration in reasoning processes~\cite{Search-o1}. (3) \textbf{Reinforcement Learning Models} R1-Gen: Pure RL-optimized generation without search engine interfacing~\cite{deepseekai2025deepseekr1incentivizingreasoningcapability}; Search-R1: Multimodal trajectory optimization combining search interactions with reasoning steps~\cite{jin2025searchr1trainingllmsreason}; ZeroSearch: Supervised LLM transformation into dual-function retrieval module (relevant/noisy document generation)~\cite{sun2025zerosearchincentivizesearchcapability}; StepSearch: PPO-based search LLM training with incremental exploration~\cite{wang2025stepsearchignitingllmssearch}. 

\textbf{Implementation Details.}
To maintain consistency with previous work, we selected NQ and HotpotQA as our training datasets. Using the data construction and evaluation frameworks outlined in Appendices~\ref{appendix:data_construction} and~\ref{appendix:data_evaluation}, we obtained a total of 71K training samples. All comparison methods use Qwen2.5-7B-Instruct as the base model. We employ E5-base-v2~\cite{wang-etal-2024-improving-text} as the retriever and utilize Wikipedia data from December 2018 as the knowledge base~\cite{karpukhin-etal-2020-dense}. All corpus indexing and embedding processes are pre-processed using FlashRAG~\cite{flashrag2025}. Except for ReSearch, which retrieves the top 5 documents for each question, all other methods retrieve the top 3 documents. For additional implementation details, please refer to Appendix~\ref{appendix:implementation_details}. \textbf{Thinker} represents Thinker model, \textbf{KAG-Thinker} represents the integration of KAG framework and Thinker model.


\subsection{Main Results}
We evaluated our model on seven widely used benchmark datasets. Here, to maintain consistency with the retriever components of other baseline methods, we conduct plain text retrieval using the \textit{"Step"} content from the logical form. Table~\ref{tab:main_results} presents the comparisons between our model and other baselines. Compared to baselines without retrieval, the performance of our model exceeds Naive Generation and CoT by an average of 27. 1\% and 34. 6\%, respectively, in seven datasets. The primary reason for the improvement is that, within these datasets, retrieval is essential, as LLMs struggle to directly answer these questions accurately without access to relevant information.
Compared to retrieval-augmented methods, our model outperforms Search-o1, IRCoT, and Naive RAG by average margins of 24.6\%, 22.6\%, and 14.8\%, respectively. These enhancements are primarily attributed to our model's implementation of breadth decomposition and depth solving, enabling it to learn how to leverage the retriever more effectively—particularly during deep retrieval, where the generated queries align well with the retriever's capabilities.
In comparison to reinforcement learning-based approaches, it can be observed that our model outperforms the previous state-of-the-art model, ReSearch, by an average of 4.1\% in EM score across seven datasets. Specifically, it improves by an average of 4.5\% on single-hop datasets and by an average of 3.9\% on multi-hop datasets. The primary reason is that our breadth decomposition and depth solving model enables the decomposition of questions into atomic granularity, thereby reducing the complexity involved in model retrieval and answering.

\renewcommand\arraystretch{1.0}
\begin{table}[ht]
\centering
\small
\setlength\aboverulesep{0pt}\setlength\belowrulesep{0pt}
\begin{tabular}{p{2.8cm}|p{1.2cm}<{\centering}p{1.2cm}<{\centering}p{1.2cm}<{\centering}|p{1.2cm}<{\centering}p{1.2cm}<{\centering}p{1.2cm}<{\centering}p{1.4cm}<{\centering}|p{1.2cm}<{\centering}}
\toprule
\multirow{2}{*}{} & \multicolumn{3}{c}{Single-Hop QA} & \multicolumn{4}{|c|}{Multi-Hop QA}       & \multirow{2}{*}{Avg} \\ \cline{2-8}
                  & NQ$^\dagger$      & TriviaQA$^\ast$   & PopQA$^\ast$   & HotpotQA$^\dagger$ & 2Wiki$^\ast$ & MuSiQue$^\ast$ & Bamboogle$^\ast$ &                      \\ \toprule
Naive Generation  & 0.134   & 0.408      & 0.140   & 0.183    & 0.250 & 0.031   & 0.120     & 0.181                \\ 
CoT               & 0.048   & 0.185      & 0.054   & 0.092    & 0.111 & 0.022   & 0.232     & 0.106                \\ \hline
Search-o1         & 0.151   & 0.443      & 0.131   & 0.187    & 0.176 & 0.058   & 0.296     & 0.206                \\
IRCoT             & 0.224   & 0.478      & 0.301   & 0.133    & 0.149 & 0.072   & 0.224     & 0.226                \\
Naive RAG         & 0.349   & 0.585      & 0.392   & 0.299    & 0.235 & 0.058   & 0.208     & 0.304                \\ \hline
R1-Gen            & 0.270   & 0.537      & 0.199   & 0.237    & 0.292 & 0.072   & 0.293     & 0.271                \\
Search-R1         & 0.393   & 0.610      & 0.397   & 0.370    & 0.414 & 0.146   & 0.368     & 0.385                \\
ZeroSearch        & 0.436   & \textbf{0.652}      & \textbf{0.488}   & 0.346    & 0.352 & 0.184   & 0.278     & 0.391                \\ \cdashline{1-9}[2pt/2pt]
StepSearch        & -       & -          & -       & 0.386    & 0.366 & \textbf{0.226}   & 0.400     & -                    \\
ReSearch          & 0.407   & 0.611      & 0.423   & 0.419    & 0.412 & 0.205   & 0.400     & 0.411                \\ \hline
Thinker (ours)        & \textbf{0.450}   & 0.642      & 0.484   & \textbf{0.421}    & \textbf{0.469} & 0.221   & \textbf{0.480}     & \textbf{0.452}                \\ \bottomrule
\end{tabular}
\caption{EM performance of different models on Qwen2.5-7B-Instruct. The best performance is set in bold. $^\dagger$/$^\ast$ represents in-domain / out-domain datasets. In contrast to other baselines, StepSearch and ReSearch employ the Musique dataset for training.}
\label{tab:main_results}
\end{table}

\subsection{Thinker with KAG}
\begin{figure}[htbp]
\centering 
\includegraphics[width=0.7
\textwidth]{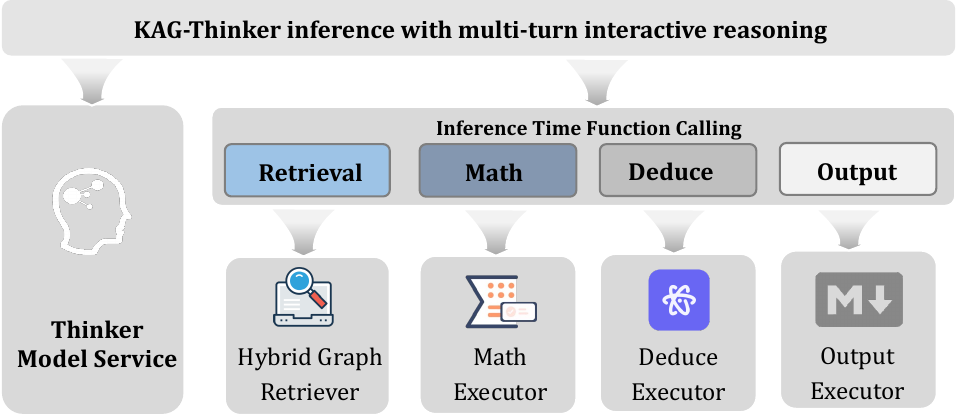} 
\caption{Overview of KAG-Thinker multi-turn interactive thinking and deep reasoning.} 
\label{Fig.thinker_with_kag} 
\end{figure}

We create KAG-Thinker by applying the Thinker model within the KAG framework. The multi-turn interactive reasoning framework of KAG-Thinker is depicted in Figure~\ref{Fig.thinker_with_kag}.
To maintain consistency with KAG~\cite{DBLP:journals/corr/abs-2409-13731}, we continue to use its test set, with 1,000 examples each from HotpotQA, Musique, and 2Wiki (consistent with HippoRAG~\cite{DBLP:conf/nips/GutierrezS0Y024}). The data for our self-built corpus are entirely derived from the supporting facts of these three datasets, totaling 26,990 documents. Within the self-built corpus, and utilizing the same retriever and retrieved documents, our Thinker model still outperforms ReSearch and Search-R1 across three multi-hop datasets.
\renewcommand\arraystretch{1.1}
\begin{table}[ht]
    \small
    \centering
\setlength\aboverulesep{0pt}\setlength\belowrulesep{0pt}
\begin{tabular}{l|c|l|cc|cc|cc|cc}
\toprule
\multirow{2}{*}{Methods}   & 
\multirow{2}{*}{Size}    & 
\multirow{2}{*}{Retriever}    & 
\multicolumn{2}{c|}{HotpotQA} &    
\multicolumn{2}{c|}{MusiQue} &   
\multicolumn{2}{c|}{2Wiki}   & 
\multicolumn{2}{c}{Avg}  \\ \cline{4-11}    
            &       &  & EM       & F1     & EM      & F1    & EM    & F1  & EM    & F1    \\
            \midrule
Search-R1 & 7B & BGE-M3
& 0.545    & 0.676 & 0.307   & 0.403  & 0.517 & 0.589  & 0.456 & 0.556  \\
ReSearch & 7B & BGE-M3
& 0.538    & 0.671 &  0.322   & 0.423   & 0.555  & 0.633   & 0.472  & 0.576 \\
Thinker (ours)  & 7B & BGE-M3
& 0.538    & 0.664 &    0.337    & 0.442    & 0.596 & 0.657   & 0.490 & 0.588  \\ \hline
KAG-Thinker (ours) & 7B  & HGR
& \textbf{0.568} & \textbf{0.698} & \textbf{0.350}   & \textbf{0.469} & \textbf{0.644} & \textbf{0.711}  & \textbf{0.520} & \textbf{0.626} \\
\bottomrule
\end{tabular}
    \caption{Performance of different models within a self-built corpus. As KAG-Thinker 7B does not possess Math and Deduce execution capabilities, we utilize the executor from KAG-V0.8 72B. All retrievers retrieve the top 3 documents.}
    \label{tab:retriever_experiments}
\end{table}
For a fair comparison against the baselines, the original Thinker model could only perform pure text retrieval with BGE-M3 for natural language steps in its logical form, even for operations like Deduce and Math. However, the KAG framework robustly supports these operations. Table~\ref{tab:retriever_experiments} illustrates that KAG-Thinker, operating within the KAG framework, shows a marked improvement over Thinker, with average EM and F1 scores increasing by 3.0\% and 3.8\%, respectively. KAG-Thinker achieves these enhancements by leveraging two key features from the KAG framework: the hybrid graph retriever (HGR) and native support for Math and Deduce. 

\subsection{Stability Analysis}
\begin{figure}[htbp]
\centering 
\includegraphics[width=\textwidth]{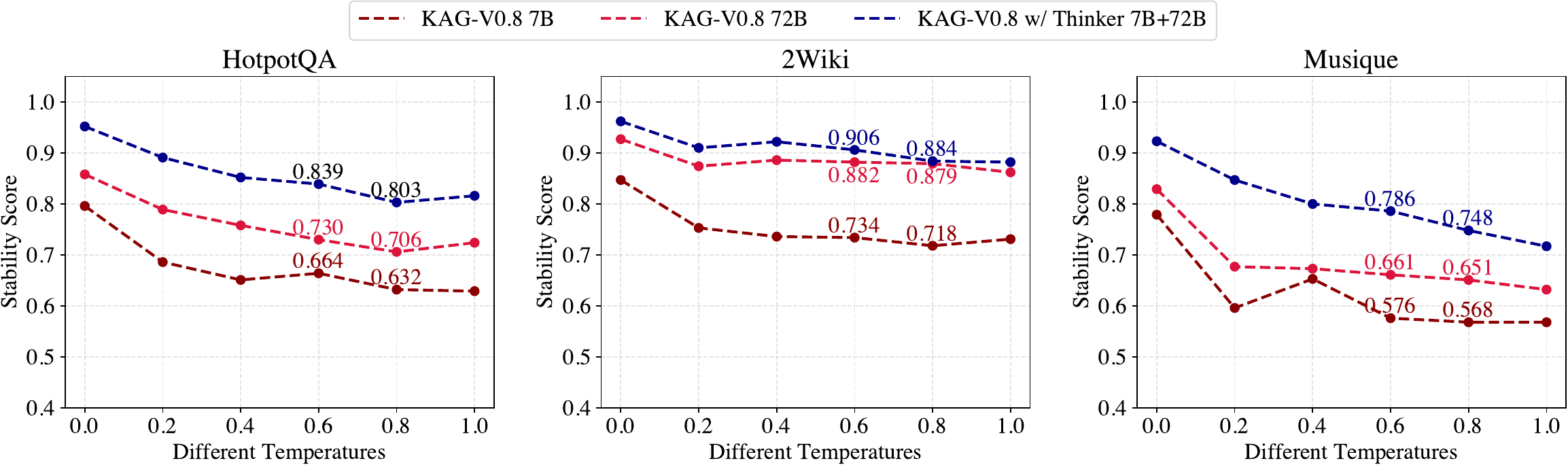} 
\caption{Stability under different temperatures.} 
\label{Fig.stability_score} 
\end{figure}

To address the inconsistency of KAG-V0.8, which often produces different solution steps on repeated attempts, we employ Thinker as the planner for KAG-V0.8. By training it on data for both breadth decomposition and depth solving, we teach the LLM to tackle problems more systematically, leading to a marked improvement in stability. For a more intuitive explanation, we introduce a stability score. For evaluation, we ask a question twice using different random seeds, thereby obtaining two distinct solution processes for each question. Then these two outputs are evaluated by LLM. If LLM determines that the two results are consistent, the question's solution steps are deemed stable; otherwise, they are considered unstable. As illustrated in Figure \ref{Fig.stability_score}, KAG-V0.8 with Thinker outperforms KAG-V0.8 7B and KAG-V0.8 72B in terms of stability on HotpotQA, 2Wiki, and Musique. Under commonly used temperature parameters of 0.6 and 0.8, KAG-V0.8 with Thinker demonstrates an average improvement of 17.9\% and 7.6\% over KAG-V0.8 7B and KAG-V0.8 72B, respectively.

\renewcommand\arraystretch{1.0}
\begin{table}[ht]
\centering
\small
\setlength\aboverulesep{0pt}\setlength\belowrulesep{0pt}\setlength{\tabcolsep}{4pt}
\begin{tabular}{l|c|ccc|ccc|ccc|ccc}
\toprule
\multirow{2}{*}{Methods} & \multirow{2}{*}{Size} & \multicolumn{3}{c|}{HotpotQA} & \multicolumn{3}{c|}{MusiQue} & \multicolumn{3}{c|}{2Wiki} & \multicolumn{3}{c}{AVG} \\ \cline{3-14}
                         &                       & EM       & F1      & LLM     & EM      & F1      & LLM     & EM      & F1     & LLM    & EM     & F1     & LLM   \\ \midrule
Naive   Gen              & 72B                   & 0.223    & 0.313   & 0.342   & 0.033   & 0.074   & 0.083   & 0.199   & 0.310  & 0.382  & 0.152  & 0.232  & 0.269 \\
Naive   RAG              & 72B                   & 0.566    & 0.704   & 0.762   & 0.248   & 0.357   & 0.384   & 0.448   & 0.512  & 0.573  & 0.421  & 0.524  & 0.573 \\ \hline
HippoRAGV2               & 72B                   & 0.557    & 0.694   & 0.807   & 0.289   & 0.404   & 0.452   & 0.542   & 0.618  & 0.684  & 0.463  & 0.572  & 0.648 \\
PIKE-RAG                 & 72B                   & 0.558          & 0.686          & 0.787          &  \underline {0.383}    &  \underline {0.498}    &  \underline {0.565}    & 0.630          & 0.720          & 0.810          & 0.524          & 0.635          & 0.721          \\
KAG-V0.8                 & 72B                   & \textbf{0.625} & \textbf{0.772} & \textbf{0.857} & \textbf{0.432} & \textbf{0.569} & \textbf{0.626} &  \underline {0.692}    &  \underline {0.784}    &  \underline {0.847}    & \textbf{0.583} & \textbf{0.708} & \textbf{0.777} \\
KAG-V0.8                 & 32B                   & 0.576          & 0.718          & 0.809          & 0.358          & 0.478          & 0.525          & \textbf{0.698} & \textbf{0.786} & \textbf{0.856} & 0.544          & 0.661          & 0.730          \\ \hline
KAG-V0.8 w/ Thinker    & 7B+72B                &  \underline {0.609}    &  \underline {0.739}    &  \underline {0.836}    & 0.365          & 0.495          & 0.561          & 0.665          & 0.755          & 0.821          &  \underline {0.546}    &  \underline {0.663}    &  \underline {0.739}   
 \\ \bottomrule 
\end{tabular}
\caption{Performance of different frameworks. The best performance is indicated in bold, and the second-best is underlined. "LLM" represents the judgment accuracy of the LLM model, which is consistent with the KAG framework. 72B refers to Qwen2.5-72B-Instruct, 32B to Qwen2.5-32B-Instruct, and 7B to the trained Thinker model.}
\label{tab:framework_experiments}
\end{table}

Table~\ref{tab:framework_experiments} shows the experimental results when the planner of KAG-V0.8 is replaced with Thinker. KAG-V0.8 with Thinker demonstrates superior average performance across three datasets compared to HippoRAGV2~\cite{gutierrez2025ragmemorynonparametriccontinual} and PIKE-RAG~\cite{wang2025pikeragspecializedknowledgerationale}. For detailed experimental settings, please refer to the KAG-V0.8 release notes. While KAG-V0.8 with Thinker significantly enhances framework stability, its average performance is lower than KAG-V0.8 72B, though it slightly surpasses KAG-V0.8 32B. This indicates that the 7B Thinker model exhibits limited problem decomposition capabilities. An analysis of bad cases reveals its struggle with complex queries. For example, "Who is the paternal grandmother of John III, Duke Of Cleves?" requires two decomposition steps (identifying John III's mother, then her mother) which the model fails to perform. This inability is primarily attributed to two factors: inconsistent decomposition of the complex natural language question and the restricted generalization capacity of the 7B model. To address these issues, our future work will involve synthesizing problem decomposition samples from structured data, thereby enhancing LLMs' ability to decompose complex problems and improving the stability of this process.

\section{Medical Field Application}
In recent years, LLMs have been widely applied in the medical field. However, their limited interpretability has hindered real-world deployment, especially for complex, multifaceted problems commonly encountered in medical consultations. For instance, disease diagnosis often involves identifying candidate diseases based on symptoms and then performing differential diagnosis by reviewing the clinical manifestations of each candidate disease—a process that benefits from decomposing the problem into simpler sub-problems. Nevertheless, most existing methods attempt direct retrieval or reasoning \cite{AlphaMed,MEDCSP,ChiMed-GPT,MEDGENIE,MKRAG}, struggling to effectively address such intricate issues .

On the other hand, for relatively straightforward medical commonsense or symptom-related queries, such as \textit{"What is the normal range for heart rate?"} or \textit{"What symptoms are associated with hypoglycemia?"}, LLMs can often generate satisfactory answers. For more rigorous questions, such as those involving drug composition or dosage analysis, LLMs are prone to hallucinations, making it necessary to retrieve external knowledge. However, existing approaches \cite{MEDGENIE,MKRAG} typically do not analyze whether retrieval is truly necessary, leading to cases where retrieval introduces additional noise, particularly when the knowledge base is incomplete. 


To address these issues, we design an interpretable model. By decomposing problems into logically dependent sub-problems, the problem-solving logic becomes more rigorous and interpretable. Through knowledge boundary determination, it reduces the introduction of unnecessary noisy references, thereby improving the accuracy of the model's answers.

\renewcommand\arraystretch{1.1}
\begin{table}[htbp]
\centering
\small
\setlength\aboverulesep{0pt}\setlength\belowrulesep{0pt}
\begin{tabular}{p{0.97\textwidth}}
\toprule
\breadthquestion{\textbf{Question}: Boy, 3 years old, weighing 16kg, sustained scald burns on both lower limbs from hot water. Physical examination: BP 85/60 mmHg, restless. Both lower limbs (including buttocks) suffered second-degree burns, with a urine output of 15ml/h. The amount of colloid fluid that should be administered in the first 24 hours is (    ). Options: A: 500ml B: 800ml C: 400ml D: 600ml E: 700ml} 
\hline
\functionfirst{\textbf{Breadth Decomposition}: 

\textbf{<think>} The user's question concerns a 3-year-old boy weighing 16kg...

First, we need to calculate the fluid volume based on the burn area and the child's weight...The proportion of colloid fluid to crystalloid fluid is typically 1:1 or 0.5:1, depending on the severity of the burn. Next, we need to clarify several key aspects: ... \textbf{</think>}

\textbf{Step 1: Fluid Requirement Formula}

Action 1: Retrieval(s=s1:Medical formula[`Fluid Requirement Formula'], p=p1, o=o1)

\textbf{Step 2: Fluid Replacement Formula for the First 24 Hours}

Action 2: Retrieval(s=s1, p=p2:time, o=o2:time[`The first 24 hours'])

\textbf{Step 3: The Percentage of Total Body Surface Area (TBSA) for Both Lower Limbs (Including the Buttocks)}

Action 3: Retrieval(s=s2:BodyParts[`Both Lower Limbs (Including the Buttocks)'], p=p3:AreaRatio, o=o3:AreaRatio)

\textbf{Step 4: The Percentage of Total Body Surface Area (TBSA) for Both Lower Limbs (Including the Buttocks) in Children}

Action 4: Retrieval(s=s2, p=p4:BelongTo, o=o4:Crowd[`Children'])

\textbf{Step 5: The Proportion of Colloid Fluid in the Total Fluid Volume for Second-Degree Burns}

Action 5: Retrieval(s=s3:TreatmentPlan[`Burn fluid replacement'], p=p5:Colloidal liquid ratio, o=o5:Colloidal liquid ratio, p.Degree of burn=II$^{\circ}$)

\textbf{Step 6: Based on \#1, \#2, \#3, \#4, and \#5, calculate the amount of colloid fluid to be administered in the first 24 hours.}

Action 6: Math(content=[`o1', `o2', `o3', `o4', `o5', `weight=16kg'], target=`The amount of colloid fluid to be administered in the first 24 hours')$\to$math6
} \hline

\functionfirst{\textbf{Depth Solving}:

\textbf{Subanswer 1}: The formula for calculating fluid replacement in burn patients is often based on the Parkland formula...

\textbf{Subanswer 2}: According to burn formulas, the basic principle for fluid resuscitation in pediatric burns is:

Total fluid volume for the first 24 hours = Body weight (kg) $\times$ Burned percentage (\%TBSA) $\times$ 1.5 (pediatric coefficient) + baseline physiological fluid requirement.

\textbf{Subanswer 3}: The burn area of both lower limbs (including the buttocks) accounts for approximately 46\% of the Total Body Surface Area (TBSA) in adults...

\textbf{Subanswer 4}: The burned area for both lower limbs (including buttocks) accounts for approximately \textcolor{red}{46\%} of TBSA in adults, but the percentage may differ for children and needs to be further verified. \textcolor{blue}{<search>Step: The Percentage of Total Body Surface Area (TBSA) for Both Lower Limbs (Including the Buttocks) in Children. Action: Retrieval(s=s2, p=p4:BelongTo, o=o4:Crowd[`Children'])</search>}

<Reference>...</Reference>

\textcolor{teal}{The area of both lower limbs (including the buttocks) for children is calculated as 46\% - (12 - age)\% = 37\%.}

\textbf{Subanswer 5}: Parkland Formula: \textcolor{red}{1ml} of fluid per kilogram of body weight per 1\% of burned area. Half of the total fluid volume is colloid fluid. However, considering the requirements for the first 24 hours and the differences in pediatric cases, I need to confirm further. \textcolor{blue}{<search>Step: The Proportion of Colloid Fluid in the Total Fluid Volume for Second-Degree Burns. Action: Retrieval(s=s3:TreatmentPlan[`Burn fluid replacement'], p=p5:Colloidal liquid ratio, o=o5:Colloidal liquid ratio, p.Degree of burn=II$^{\circ}$)</search>}

<Reference>...</Reference>

\textcolor{teal}{According to the 7th Edition of the Surgery Fluid Resuscitation Protocol, the amount of colloid and electrolyte solution to be provided for pediatric patients is 2.0ml, with a ratio of 1:1 between colloid and electrolyte solution.}

\textbf{Subanswer 6}: <Python>...</Python>

\textcolor{teal}{592ml}
} \hline
\breadthanswer{\textbf{Answer}:

1.Determine Burn Area:

The patient is ..., the total burn area is calculated to be 37\%.

2.Calculate Colloid Volume:

According to the fluid replacement protocol ... Colloid volume = Body weight (kg) × Total burn area (\%) × 1
= 16 kg × 37\% × 1 = 592 ml.

Since 592 ml is approximately 600 ml, option D is selected.
}
\bottomrule
\end{tabular}
\caption{An example of the problem-solving process in KAG-Med-Thinker: The breadth decomposition module first generates the thought process, then decomposes the problem into multiple sub-problems. In the depth solving module, an adaptive retrieval equilibrium model is used for reasoning and external knowledge retrieval. Finally, the answer is generated based on the context. The red font part represents errors in the questions generated by the model, the blue font part represents adaptive generation of retrieval instructions, and the green font part represents the results executed by logical form executors.}
\label{tab:med_planner}
\end{table}

\subsection{Medical Thinker Model}
To address the key points mentioned above, within the framework of KAG-Thinker, we use a breadth decomposition approach to decompose the problem into multiple sub-problems and their corresponding logical form, and we also employ knowledge boundary model to determine whether retrieval is necessary. Table \ref{tab:med_planner} illustrates the reasoning process of KAG-Med-Thinker. First, within the breadth decomposition part, the content between \textcolor{blue}{<think>} and \textcolor{blue}{</think>} represents the thought process. Here, it is reasoned that calculating \textit{the required fluid volume} involves considering both \textit{the burn area and weight}. Consequently, information regarding \textit{the percentage of lower limb (including buttock) surface area in children}, as well as \textit{the ratio of colloidal fluids to crystalloid fluids}, should be retrieved for further decomposition of the sub-problems. Secondly, in the depth solving part, each sub-problem is addressed independently. For sub-problems related to retrieval, the system first performs its own reasoning and then evaluates whether the answer is reliable and complete. If not, it generates a retrieval directive using <search>...</search>. The Chunks retrieved for each sub-problem (with top-k=3) are integrated into the context. Since the retrieved Chunks may be lengthy, we perform knowledge compression and information targeting on the results for each sub-problem. The answers are then refined on the basis of the reference. For sub-problems requiring retrieval (e.g., Step 4 and Step 5), a deep retrieval check is employed to assess whether the Reference can resolve the sub-problem. If the Reference cannot address the sub-problem, a new retrieval directive is generated. Specifically, for Step 6, the Math directive is executed by generating Python code and obtaining computational results using a code interpreter. Finally, based on the breadth decomposition and depth solving, the complete reasoning process and the final result are generated.

We structure the samples into a multi-turn interactive reasoning format. For synthesizing data, question decomposition is performed using Qwen2.5-72B-Instruct, which offers stronger reasoning capabilities and instruction-following abilities, allowing distillation into smaller models. The target model is trained to generate answers for sub-problems and determine whether external executors are required. Since the medical Q\&A often involves lengthy answers for sub-problems, we call the LLM twice to confirm whether the answers are correct and complete, which serve as labels for the knowledge boundary determination. The subsequent deep retrieval, reasoning, and answer synthesis utilize distilled synthetic data derived from the LLMs.

\subsection{Medical Knowledge Base}
\subsubsection{Medical Data Sources}
We construct a Chinese medical knowledge retrieval repository containing data sourced from authoritative medical textbooks and professional medical websites. The \textbf{Medical Textbooks} originate from the Chinese materials in MedQA, specifically related to resources from the National Medical Licensing Examinations in China (e.g., the Licensed Physician Qualification Examination). These materials are highly professional and practical, making them well-suited for research and development of medical intelligent question-answering systems tailored for Chinese users. \textbf{Professional Medical Websites}. Medical data were primarily collected from five professional medical websites: Xiahe Medical Encyclopedia, DXY Doctor, Tencent Medical Encyclopedia, Kuaishou Medical Encyclopedia, and Baidu Health.

\begin{figure}[htbp]
\centering 
\includegraphics[width=\textwidth]{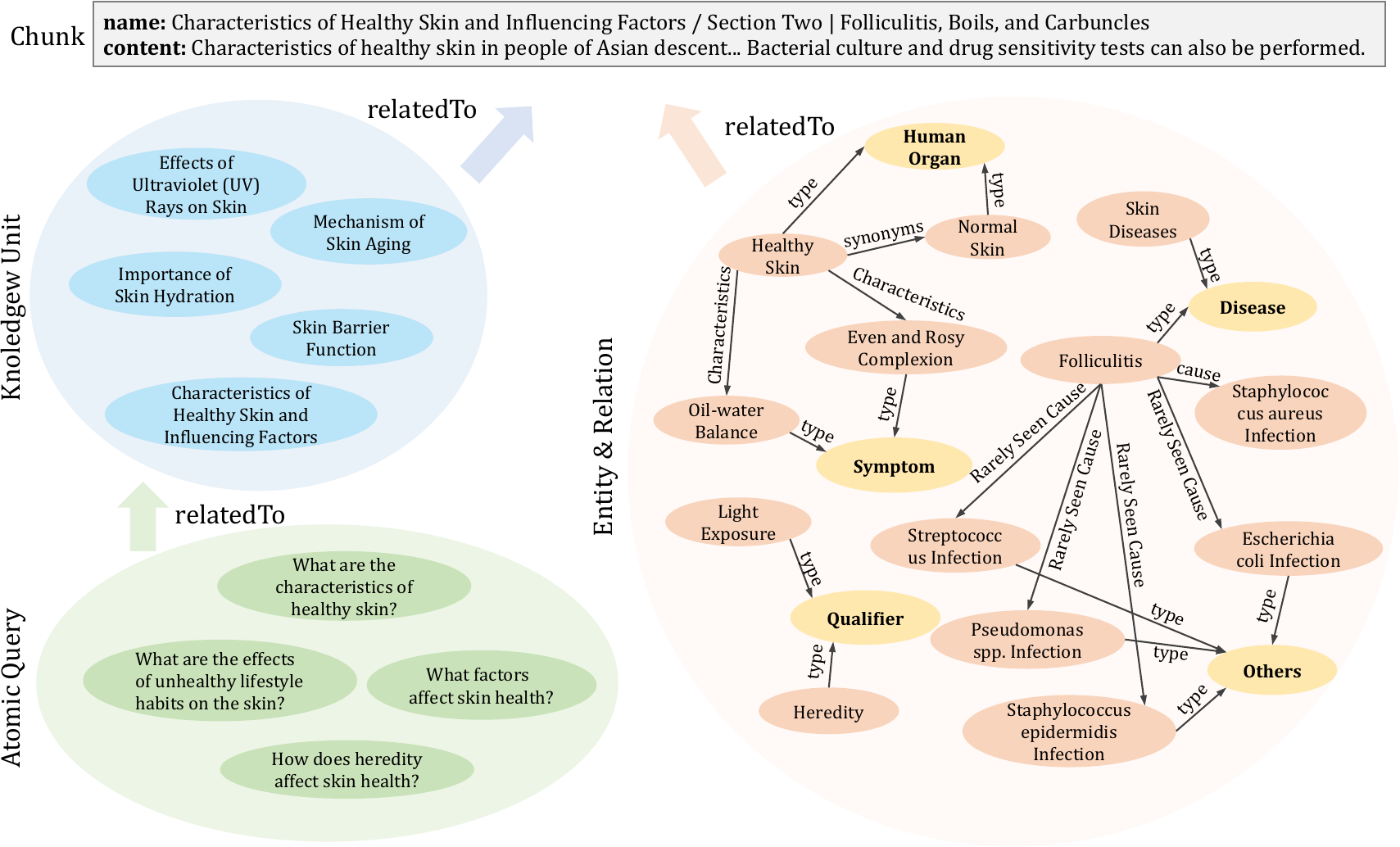} 
\caption{Medical domain knowledge base: Knowledge consists of Entities and Relations, Knowledge Units, and Atomic Query nodes. The Entities and Relations, as well as the Knowledge Units, are extracted from the original chunks. Relevant Atomic Query nodes are then generated based on the Knowledge Units. These components are interconnected and linked to the chunks.} 
\label{Fig.med-KG} 
\end{figure}

\subsubsection{Medical Knowledge Base Construction}
\label{sec.medical_knowledge_graph}

For the extraction of information from the medical domain, we adopt schema constraint extraction due to the low error tolerance in the medical context. The use of a predefined schema restricts the extraction and generation to specific, structured key information (e.g., disease names, drug dosages, treatment plans), avoiding the introduction of irrelevant or erroneous content. We select 23 commonly used medical entity types that cover the main semantic categories involved in clinical diagnosis and treatment (e.g., diseases, symptoms, medications, diagnostic tests) and extend to general concepts related to the medical domain, public health, and system integration (e.g., disciplines, organizations, biological processes). This classification system refers to international medical terminology standards (for example, SNOMED CT, ICD-10) and ontologies of representative medical knowledges, which effectively support subsequent structured knowledge fusion, reasoning, and generation tasks. Specific medical entity types can be found in Appendix ~\ref{appendix:med_entities}. The detailed structured knowledge extraction process and prompt references can be found in Appendix ~\ref{appendix:med_knowledge_graph_extraction}. Figure ~\ref{Fig.med-KG} illustrates the constructed medical knowledge base and shows the connectivity of the structured example information.

\subsection{Experimental Settings and Results}

\subsubsection{Experimental Settings}

\textbf{Dataset}. We selected MedQA~\cite{DBLP:journals/corr/abs-2009-13081} as the evaluation dataset. This dataset is a collection of question-and-answer pairs sourced from professional medical examinations. All questions are multiple-choice, covering areas such as medical knowledge, disease diagnosis, pharmacology, and diagnostic procedures. Since the original test dataset is very large, we randomly sampled 300 entries from the Chinese subset to form our evaluation dataset. 

\textbf{Implementation Details}. We conducted SFT (Supervised Fine-Tuning) training on three base models, Meta-LLAMA-3.1-8B-Instruct, Qwen2.5-14B-Instruct and DeepSeek-R1-Distill-Qwen-14B, with the following configurations: learning rate lr=5e-6, epoch=5, and using a self-constructed medical KB in Section~\ref{sec.medical_knowledge_graph}, while leveraging BGE-M3 to build vector indexes.


\subsubsection{Main Results}

The empirical evaluation results are presented in Table \ref{tab:med-result}. KAG-Med-Thinker consistently demonstrates enhanced performance across both base models, with a particularly significant improvement being observed on the Meta-LLAMA-3.1-8B-Instruct. In particular, compared to the established multi-turn planning and retrieval augmented models, IRCoT and ReAct, KAG-Med-Thinker achieves remarkable performance improvements of 11.33\%, 34.78\% respectively. It also outperforms the Naive RAG, by a margin of 12.46\%. Among them, ReAct struggles to follow the instruction format on LLama3, resulting in poorer performance. On the DeepSeek-R1-Distill-Qwen-14B, KAG-Med-Thinker shows similar superiority, outperforming IRCoT, ReAct, and adaptive Naive RAG model by 3.95\%, 4.41\%, and 3.8\%, respectively. The baselines based on Qwen2.5-14B-Instruct have already achieved good performance, but KAG-Med-Thinker still outperforms other reasoning and RAG models, achieving the best results. These consistent gains across different base models and comparative baselines unequivocally validate the effectiveness and robustness of KAG-Med-Thinker.


\renewcommand\arraystretch{1.1}
\begin{table}[ht]
\centering
\small
\setlength\aboverulesep{0pt}\setlength\belowrulesep{0pt}
\begin{tabular}{p{4.5cm}|p{6.0cm}|p{2.0cm}<{\centering}}
\toprule
LLM & Methods & MedQA   \\ \midrule
\multirow{5}{*}{Meta-LLAMA-3.1-8B-Instruct}  & Naive Generation~\cite{qwen2025qwen25technicalreport} & 58.33  \\ 
    & Naive RAG~\cite{DBLP:conf/nips/LewisPPPKGKLYR020} & 61.54  \\ 
    & Naive RAG+adaptive~\cite{DBLP:conf/naacl/JeongBCHP24}   & 61.33   \\ 
    & IRCoT~\cite{DBLP:conf/acl/TrivediBKS23}   & 62.67   \\ 
    & ReAct~\cite{DBLP:conf/iclr/YaoZYDSN023}   & 39.22   \\ 
    & \textbf{KAG-Med-Thinker}   & \textbf{74.00}  \\
\hline
\multirow{5}{*}{Qwen2.5-14B-Instruct}  & Naive Generation~\cite{qwen2025qwen25technicalreport} & 84.00  \\ 
    & Naive RAG~\cite{DBLP:conf/nips/LewisPPPKGKLYR020} & 81.67  \\ 
    & Naive RAG+adaptive~\cite{DBLP:conf/naacl/JeongBCHP24}   & 85.00   \\ 
    & IRCoT~\cite{DBLP:conf/acl/TrivediBKS23}   & 82.33   \\ 
    & ReAct~\cite{DBLP:conf/iclr/YaoZYDSN023}   & 86.20   \\ 
    & \textbf{KAG-Med-Thinker}   & \textbf{87.00}  \\ \hline
\multirow{5}{*}{DeepSeek-R1-Distill-Qwen-14B}  & Naive Generation~\cite{deepseekai2025deepseekr1incentivizingreasoningcapability} & 79.67  \\ 
    & Naive RAG~\cite{DBLP:conf/nips/LewisPPPKGKLYR020} & 79.00  \\ 
    & Naive RAG+adaptive~\cite{DBLP:conf/naacl/JeongBCHP24}   & 81.48  \\ 
    & IRCoT~\cite{DBLP:conf/acl/TrivediBKS23}   & 81.33   \\ 
    & ReAct~\cite{DBLP:conf/iclr/YaoZYDSN023}   & 80.87   \\ 
    & \textbf{KAG-Med-Thinker}   & \textbf{85.28}  \\
\bottomrule
\end{tabular}
\caption{Accuracy of different models on the MedQA dataset. The best performance is set in bold.}
\label{tab:med-result}
\end{table}


Furthermore, it is observed that the adaptive Naive RAG consistently outperforms its exhaustive retrieval-based counterpart. In particular, it demonstrates a significant performance advantage of 3.33\% on Qwen2.5-14B-Instruct and 2.48\% on DeepSeek-R1-Distill-Qwen-14B. This superior performance  even extends to IRCoT and ReAct, models that employ multi-turn retrieval planning. This finding shows that unconstrained or unfiltered retrieval introduces inherent limitations and noise that negatively impact overall reasoning performance. Consequently, it is essential to judiciously determine the model's intrinsic knowledge boundaries and synergistically integrate its internal knowledge with external tool invocation to fully harness their complementary strengths.



\section{Conclusion}
KAG-Thinker is presented as a novel, cognitively-inspired reasoning framework engineered to address the limitations of extant methodologies in solving intricate multi-hop problems demanding rigorous logical coherence and contextual consistency. The framework's core methodology emulates human cognitive processes by decomposing complex queries into logically interconnected sub-problems through a structured logical-form-based reasoning mechanism, thereby enabling incremental problem resolution while preserving structural integrity. Each logical-form task is associated with a unique logical function representation and can be solved independently. This design enables the effective decoupling of knowledge retrieval and reasoning analysis into separate subtasks. Consequently, the retrieval component can be seamlessly integrated with an automated knowledge boundary detection module, which optimizes efficiency by intelligently arbitrating between internal knowledge utilization and external information retrieval, thereby enabling the LLM to decide whether to retrieve external information based on its internal assessment. Furthermore, an integrated focusing-and-reasoning module refines retrieved content and synergizes external data with internal inferential processes, mirroring human iterative knowledge refinement. The applicability of KAG-Thinker in high-stakes domains is empirically validated via a medical Q\&A system, showcasing its capacity to integrate domain-specific knowledge while maintaining logical rigor.

\section{Contributors}
\label{sec:contributors}

Authors are listed \textbf{alphabetically by the first name}.

\begin{multicols}{4}
\raggedcolumns
Dalong Zhang \\
Jun Xu \\
Jun Zhou \\
Lei Liang \\
Lin Yuan \\
Ling Zhong \\
Mengshu Sun \\
Peilong Zhao \\
QiWei Wang \\
Xiaorui Wang \\
Xinkai Du \\
YangYang Hou \\
Yu Ao \\
ZhaoYang Wang \\
Zhengke Gui \\
ZhiYing Yi \\
Zhongpu Bo
\end{multicols}

\textbf{Special acknowledgments}.

Haofen Wang@Tongji University \\
Huajun Chen@Zhejiang University \\
Zhejiang University-Ant Group KG Joint Laboratory

\bibliographystyle{assets/plainnat}
\bibliography{main}

\newpage
\beginappendix
\section{Logical Form}
\label{appendix:logical_form}
Logical Functions are defined as Table~\ref{tab:function-syntax}, with each function representing an execution action. Complex problems are decomposed by planning a combination of these expressions, enabling reasoning about intricate issues.
\renewcommand\arraystretch{1.0}
\begin{table*}[ht]
    \small
    \centering
    \setlength\aboverulesep{0pt}\setlength\belowrulesep{0pt}
    \begin{tabular}[c]{p{1.4 cm}|p{7 cm}|p{6.5 cm}}
       \toprule
       Function & Function Expression & Description \\
       \cline{1-3} 
       \midrule
       \textit{Retrieval}  
       & $Retrieval$($s$ = $s_i$:type[name], $p$ = $p_i$:edge, $o$ = $o_i$:type[name], $s$.prop = value, $p$.prop = value, $o$.prop = value) & Retrieve <$s, p, o$> triples. Conditions can be set as constraints, and $s_i, p_i, o_i$ serve as variable names for reference in subsequent action planning. When referring to previously mentioned variable names, there is no need to regenerate entity types and names. \\
       \midrule
       \textit{Deduce} 
       & $Deduce$(\texttt{op} = extract | judgement | entailment | choice,        \texttt{content} = [A,B,...], \texttt{target} = [...]) $\rightarrow$ $deduce_i$ &  The parameter \texttt{op} can specify different inference tasks, such as information extraction, judgment, logical reasoning, single-choice, multiple-choice, etc. The parameter \texttt{content} specifies contextual information and can be either texts or variables. The \texttt{target} specifies the objective of the inference. \\
       \midrule
       \textit{Math} & $Math$ (\texttt{content} = [A,B,...], \texttt{target} = [...]) $\rightarrow$ $math_i$ &  Numerical or statistical calculations can be performed. The \texttt{content} specifies contextual information and can be either text or variables. The \texttt{target} specifies the calculation objective. \\
       \midrule
       \textit{Output} & $Output$ (A,B,...) & Directly output A, B, ... as the answer, where A and B are variable names referring to the results of previous retrieval, deduce, or calculation. \\
       \bottomrule
    \end{tabular}
    \caption{Different functions of logical form.}
    \label{tab:function-syntax}
\end{table*}

\section{Data Construction Framework}
\label{appendix:data_construction}
\begin{figure}[htbp]
\centering 
\includegraphics[width=0.5
\textwidth]{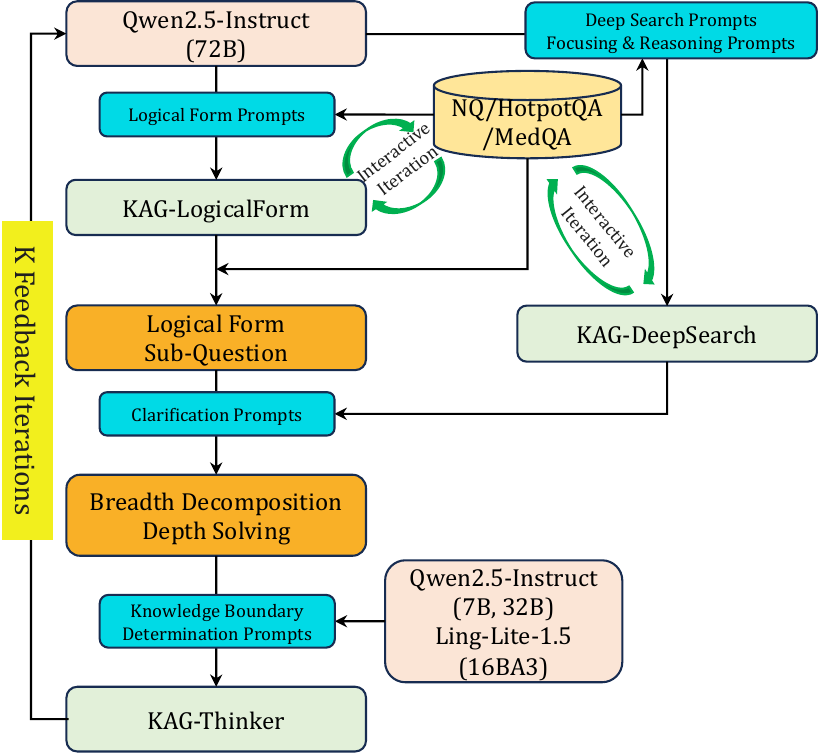} 
\caption{Overview of the data construction framework.} 
\label{Fig.dataset_construction} 
\end{figure}

RL based methods often struggle with controlling their outputs in a clear and organized way. They can't effectively customize the intermediate steps of reasoning while making decisions. Instead, these methods depend on random exploration within LLMs, leading to unpredictable reasoning paths. This unpredictability is especially problematic in important professional fields that need clear steps and the ability to review processes, such as: medical diagnostics, legal compliance, and financial risk assessment. To address these limitations, we have developed a systematic Search QA synthesis framework with customizable reasoning process specification, as depicted in Figure~\ref{Fig.dataset_construction}. Our data synthesis primarily covers the general and medical domains. For the general domain, a series of logical form prompts is designed (refer to Appendix~\ref{appendix:logical_form}). Using Qwen2.5-72B-Instruct, we perform inference on the NQ and HotpotQA datasets to generate natural language to logical form training data. This data is then filtered for correctness using a data evaluation framework (refer to Appendix~\ref{appendix:data_evaluation}). Subsequently, the first version of the KAG-LogicalForm model is trained. This model is then used to perform inference on the NQ and HotpotQA datasets. After filtering with the data evaluation framework and multiple rounds of iteration, the final KAG-LogicalForm training data, comprising 59K training samples, is obtained. Simultaneously, we also design a series of deep search and focusing \& reasoning prompts (refer to Sec~\ref{sec:focusing_reasoning}). We perform inference on single-hop questions within the NQ and HotpotQA datasets using Qwen2.5-72B-Instruct, which leads to a sub-question deep search model. After multiple rounds of iteration, the final KAG-DeepSearch training data, comprising 18K training samples, is obtained. Then, based on the KAG-LogicalForm and KAG-DeepSearch models, we obtain the breadth decomposition and deep solving process for each question. Finally, we perform a knowledge boundary determination for each sub-question. For sub-questions that LLMs can confidently answer, we omit their deep solving process to reduce unnecessary retrieval by the model. After multiple rounds of iteration and validation by the data evaluation framework, we obtain the final KAG-Thinker training data, comprising 71K training samples. The data construction method in the medical domain is the same as that in the general domain. The training data constructed for KAG-LogicalForm, KAG-DeepSearch, and KAG-Med-Thinker amount to 14K, 1K, and 20K entries, respectively.

\section{Data Evaluation Framework}
\label{appendix:data_evaluation}

The Data Evaluation Framework (DEF) constitutes a systematic methodology for optimal dataset curation through multi-dimensional quality assessment. In contemporary machine learning paradigms, empirical evidence consistently demonstrates that data quality exerts a more critical influence on model performance than dataset scale-a particularly salient consideration given the pervasive challenge of quality inconsistencies in publicly available corpora. Motivated by this critical need, we have engineered an evaluative architecture capable of granular quality diagnostics at the query-answer pair level. the DEF operates through six hierarchically organized assessment dimensions: security, accuracy, relevance, logic, fluency, and emotion.

\textbf{Security.}
A security assessment identifies issues within data to ensure its safety and integrity. It checks if model outputs align with social norms, laws, and ethical standards. Key areas of focus include: (1) Content: Checking for sensitive, discriminatory, or inappropriate material to prevent harm.
(2) Privacy \& Security: Ensuring compliance with privacy laws and data security rules, such as preventing unauthorized access.
(3) Risk: Identifying and removing statements that could cause controversy or significant risks, which helps maintain trust and reputation.
(4)Consequences: Evaluating if actions based on the data might lead to negative outcomes.
In essence, this systematic check ensures data is accurate, legal, and ethical. It finds and resolves potential threats, fostering trust and security.

\textbf{Accuracy.}
Accuracy verification is a key process that identifies errors and inconsistencies in data and model outputs. It ensures model responses accurately address user questions and are factually sound. Key aspects of this evaluation include:
(1) Factual Error Detection: We rigorously check for any incorrect information, such as wrong dates, data, or logical flaws, to ensure reliability.
(2) User Intent Understanding: We verify that the model correctly interprets the user's question and its context, preventing irrelevant or misleading answers.
(3) Common Sense \& Domain Alignment: Outputs are checked against established knowledge and common sense principles for coherence and practical value.
(4) Illusion Detection: We look for answers that might seem correct but could actually mislead the user.
(5) Avoiding Over-Reasoning: We ensure the model doesn't provide excessive details or go beyond the scope of the question, keeping responses focused.
(6) Rejection Criteria: We define when an answer must be rejected due to irrelevance, inaccuracy, or confusion.
(7) Process Review: We examine the entire process of generating an answer for any factual errors introduced at any stage.
Crucially, judgments about factual correctness are based on the common sense embedded in the question itself. While external information can be helpful, the primary focus is on the question's inherent context and the user's expectations.

\textbf{Relevance.}
Relevance assessment evaluates whether model outputs are on-topic and appropriate for the given context. This process primarily checks how well the model's responses align with the input questions or cues. Key aspects of this evaluation include:
(1) Topic Deviation: We identify if the output strays from the main topic or includes information not directly relevant to the user's query. This ensures the information stays focused.
(2) Core Issue Identification: We assess the model's ability to pinpoint the main point of a question and provide specific answers that directly address it.
(3) Clarity and Conciseness: We ensure responses are not vague, too general, or overly detailed, which can obscure clarity. The goal is to provide concise, useful information without unnecessary embellishment.
In short, a relevance check is a systematic way to ensure model outputs are coherent and contextually appropriate. By identifying and fixing irrelevant content, we guarantee the information is not only accurate but also directly pertinent to the user's specific needs. This significantly improves the system's effectiveness and user satisfaction.

\textbf{Logic.}
Logical integrity assessment ensures that data outputs are internally consistent and free from errors. It verifies that model responses are rational and do not contain contradictions. Key areas of focus include:
(1) Logical Flow: Checking that timelines, sequences, and cause-and-effect relationships within the output are sound and consistent.
(2) Reasoning Soundness: Identifying any gaps in logical reasoning, flawed arguments, or incorrect deductions that undermine the conclusions.
(3) Real-World Alignment: Verifying that outputs align with established facts, common sense, and scientific principles governing real-world phenomena.
Ultimately, this meticulous process identifies and corrects logical flaws, thereby strengthening the reliability of the information and fostering confidence in data-driven decisions.

\textbf{Fluency.}
This assessment evaluates the language quality of model-generated text, focusing on its naturalness, coherence, and readability. Key aspects examined include:
(1) Structural Clarity and Grammar: We verify clear sentence structures and grammatical correctness to ensure easy comprehension.
(2) Conciseness: We aim to eliminate repetition, wordiness, and overly complex phrasing for clearer and more direct communication.
(3) Idiomatic Usage and Style: We ensure the language sounds natural, adhering to human-like idioms and appropriate stylistic conventions.
In essence, this assessment maintains high standards for linguistic excellence in model outputs. By systematically identifying and correcting fluency issues, we enhance overall text quality and readability, fostering effective communication and building trust.

\textbf{Emotion.}
This assessment checks the emotional and stylistic tone of model-generated outputs. Its main goal is to ensure the tone is appropriate for the given context and user. Key evaluation points include:
(1) Matching User Expectations: We ensure the output's tone (e.g., formal, friendly, humorous) aligns with what the user anticipates or what the situation demands.
(2) Avoiding Inappropriate Expressions: We prevent the model from using blunt, indifferent, or offensive tones that could alienate or confuse users.
(3) Contextual Adaptability: We verify the model can adjust its tone based on different contexts and various user preferences.
In short, this assessment ensures the model communicates with emotional intelligence and the right tone. By identifying and correcting tonal issues, it fosters more meaningful interactions, improves user engagement, and builds trust in the communication.

To validate the performance of our data evaluation framework, we select positive and negative samples from the medical and legal domains for human evaluation.
The evaluation metrics for positive samples are systematically presented in Table~\ref{tab:def_pos}. Our DEF demonstrates robust performance across individual assessment dimensions, achieving accuracy rates exceeding 90\% in all criteria. Domain-specific composite evaluations yield 88.8\% accuracy in medical and 93.8\% in legal scenarios. Crucially, the framework exhibits controlled false positive rates during positive sample validation, indicating effective preservation of valid instances. The evaluation metrics for negative samples are comprehensively detailed in Table~\ref{tab:def_neg}. Our DEF achieves classification accuracy surpassing 80\% across all individual assessment criteria, with domain-specific composite accuracy reaching 88.6\% in medical and 85.2\% in legal. Through continuous optimization of the DEF, the quality of our data has been steadily improving.
\renewcommand\arraystretch{1.0}
\begin{table}[htbp]
\centering
\small
\setlength\aboverulesep{0pt}\setlength\belowrulesep{0pt}
\begin{tabular}{p{2cm}|p{1.5cm}<{\centering}p{1.5cm}<{\centering}p{1.5cm}<{\centering}|p{1.5cm}<{\centering}p{1.5cm}<{\centering}p{1.5cm}<{\centering}}
\toprule
          & \multicolumn{3}{c|}{Medical}    & \multicolumn{3}{c}{Legal}        \\ \cline{2-7}
          & Good Case  & Bad Case  & Acc   & Good Case  & Bad Case  & Acc   \\ \hline
Security  & 72         & 0         & 1.000 & 96         & 0         & 1.000 \\
Accuracy  & 66         & 6         & 0.917 & 91         & 5         & 0.948 \\
Relevance & 70         & 2         & 0.972 & 94         & 2         & 0.979 \\
Logic     & 70         & 2         & 0.972 & 96         & 0         & 1.000 \\
Fluency   & 71         & 1         & 0.986 & 96         & 0         & 1.000 \\
Emotion   & 70         & 2         & 0.972 & 95         & 1         & 0.990 \\ 
All       & 64         & 8         & 0.888 & 90         & 6         & 0.938 \\ \bottomrule
\end{tabular}
\caption{The data evaluation framework's performance in positive cases in medical and legal fields.}
\label{tab:def_pos}
\end{table}
\renewcommand\arraystretch{1.0}
\begin{table}[htbp]
\centering
\small
\setlength\aboverulesep{0pt}\setlength\belowrulesep{0pt}
\begin{tabular}{p{2cm}|p{1.5cm}<{\centering}p{1.5cm}<{\centering}p{1.5cm}<{\centering}|p{1.5cm}<{\centering}p{1.5cm}<{\centering}p{1.5cm}<{\centering}}
\toprule
          & \multicolumn{3}{c|}{Medical}    & \multicolumn{3}{c}{Legal}        \\ \cline{2-7}
          & Good Case  & Bad Case  & Acc   & Good Case  & Bad Case  & Acc   \\ \hline
Security  & 24         & 5         & 0.828 & 36         & 9         & 0.800  \\
Accuracy  & 27         & 2         & 0.931 & 58         & 8         & 0.879  \\
Relevance & 24         & 6         & 0.800 & 42         & 8         & 0.840  \\
Logic     & 25         & 6         & 0.806 & 43         & 7         & 0.860  \\
Fluency   & 24         & 5         & 0.828 & 50         & 4         & 0.926  \\
Emotion   & 31         & 4         & 0.886 & 46         & 8         & 0.852  \\ \bottomrule
\end{tabular}
\caption{The data evaluation framework's performance in negative cases in medical and legal fields.}
\label{tab:def_neg}
\end{table}

\section{Implementation Details}
\label{appendix:implementation_details}
For training, we combine the NQ and HotpotQA training sets to create a unified dataset for KAGThinker. Training is performed on 8$\times$2 Nvidia H100 GPUs, utilizing full parameter optimization and gradient checkpointing. Key parameter settings are detailed in Table~\ref{tab.training_parameters}.
\renewcommand\arraystretch{1.0}
\begin{table}[htbp]
\centering
\small
\setlength\aboverulesep{0pt}\setlength\belowrulesep{0pt}
\begin{tabular}{p{3.5cm}p{1.5cm}p{2.5cm}p{1.5cm}}
\toprule
Parameter                 & Value    & Parameter     & Value  \\ \hline
Learning Rate             & 5e-06    & Weight Decay  & 0.0    \\
Training Batch Size       & 64       & Warmup Ratio  & 0.06   \\
Training Epochs           & 5        & Gradient Clip & 1.0    \\
Max Sequence Length       & 16384    & Seed          & 1024   \\
Optimizer                 & Adamw    & LR Scheduler  & linear \\ \bottomrule
\end{tabular}
\caption{Some important parameters for model training.}
\label{tab.training_parameters}
\end{table}

\section{Training Sample}
To more intuitively illustrate our training process, we provide a detailed example below.
\begin{minted}
[
frame=lines,
framesep=2mm,
baselinestretch=1.0,
breaklines=true,
bgcolor=LightGray,
fontsize=\footnotesize,
]
{json}
[{
	"role": "system",
	"content": "As you answer each question, you must provide a thought process and insert it between < think > and </think>."
}, {
	"role": "user",
	"content": "You are an expert in function calls, capable of accurately understanding function definitions and precisely decompose user queries to select...Question:\nwho does the honey badger play for in the nfl\nOutput:\n"
}, {
	"role": "assistant",
	"content": "<think>This is a sports inquiry problem. Our investigation ...</think>\n\n<answer>Step1: who does the honey badger play for in the nfl\nAction1: Retrieval(s=s1:player[`Honey Badger`], p=p1:playsFor, o=o1:team)\nStep2: Output #1\nAction2: Output(o1)</answer>"
}, {
	"role": "user",
	"content": "Can you answer the following questions step by step? If you can, wrap your answer with <answer>\\boxed{your answer}</answer>. If you can't, just reply that based on my internal knowledge, I can't answer this question, I need to retrieve external knowledge. \nQuestion: \nStep1: who does the honey badger play for in the nfl\nAction1: Retrieval(s=s1:player[`Honey Badger`], p=p1:playsFor, o=o1:team)"
}, {
	"role": "assistant",
	"content": "<think>based on my internal knowledge, I can't answer this question, I need to retrieve external knowledge.</think>\n<search>Step1: who does the honey badger play for in the nfl\nAction1: Retrieval(s=s1:player[`Honey Badger`], p=p1:playsFor, o=o1:team)</search>"
}, {
	"role": "user",
	"content": "<references>[0]\"Tyrann Mathieu\"\nTyrann Mathieu Tyrann Devine Mathieu (; born May 13, 1992) is an American football safety for the Houston Texans of the National Football League (NFL)...</references>"
}, {
	"role": "assistant",
	"content": "<think> Based on the search results, the nickname \"Honey Badger\" is associated with Tyrann Mathieu, an American football player. I will now search for the team he plays for.</think>\n<search>Step1:  Who does Tyrann Mathieu play for in the NFL? \nAction1:Retrieval(s=s1:human[`Tyrann Mathieu`], p=p1:playsFor, o=o1:sportsTeam)</search>"
}, {
	"role": "user",
	"content": "<references>[0]\"Tyrann Mathieu\"\nthe game on a Phil Dawson field goal. On November 26, Mathieu recorded his second Interception of the season in a 27-24 overtime win against the Jacksonville Jaguars...</references>"
}, {
	"role": "assistant",
	"content": "<think> The search results indicate that Tyrann Mathieu is currently playing for the Houston Texans in the NFL.</think>\n<answer> The Honey Badger plays for \\[ \\boxed{Houston Texans} \\] in the NFL. </answer>."
}, {
	"role": "user",
	"content": "Answer the last question based on question, the first n sub-questions and their answers (indicated by #n), and wrap the result with <answer>\\boxed{your answer}</answer>. \nquestion:\nwho does the honey badger play for in the nfl\nsub-questions:\nwho does the honey badger play for in the nfl\n#1: Houston Texans\n\nlast-question:\nOutput #1\nanswer:\n"
}, {
	"role": "assistant",
	"content": "<think>I think I can answer this question now</think>\n\n<answer>\\boxed{Houston Texans}</answer>."
}]
\end{minted}

\section{Med Entities}
\label{appendix:med_entities}

In the medical field, fully open-ended entity relationship extraction can result in a large amount of useless information, leading to unnecessary resource consumption and providing no benefit to retrieval. Therefore, we imposed constraints on the types of entities to be extracted. Based on the application scenarios, we identified the following 23 types of entities and only extracted relationships among these entities, with no constraints on the types of relationships. For details, see Table \ref{tab:entity_types}.

\renewcommand\arraystretch{1.0}
\begin{table}[ht]
  \centering
  \small
  \setlength\aboverulesep{0pt}\setlength\belowrulesep{0pt}
\begin{tabular}{l|l}
\toprule
\textbf{Entities}          & \textbf{Describe}     \\ \midrule
Disease                    & Names of various diseases, such as "diabetes", "hypertension", etc                         \\ 
Symptom                    & Subjective feelings or clinical manifestations of patients, such as "fever" and "headache" \\ 
SurgicalProcedure & Various surgical procedures and procedures, such as "cholecystectomy"                               \\ 
Drug              & Chemicals used for treating, preventing, or diagnosing diseases, such as aspirin           \\ 
Vaccine                    & Biological products for immunization, such as "hepatitis B vaccine"                                 \\ 
Examination                & Clinical physical examination items, such as "electrocardiogram examination"                        \\
LaboratoryTest             & Laboratory testing items, such as "blood routine" and "urine protein testing"                       \\
Population                 & Specific population descriptions, such as "elderly population" and "pregnant women"                 \\
MedicalDiscipline          & Medical branch disciplines, such as "Internal Medicine" and "Surgery"                               \\
Department                 & Diagnosis and treatment departments in medical institutions, such as "Cardiovascular Medicine"      \\
HumanOrgan                 & Organs in human anatomy, such as the liver and heart                                                \\
MedicalDevice              & Medical equipment for clinical use, such as "ventilators" and "electrocardiogram monitors"          \\
Organization               & Medical related units or organizations, such as "hospitals" and "disease control centers"           \\
Location                   & Geographical locations related to medical behavior, such as "outpatient department" and "ward"      \\
Gene                       & Genetic information related to diseases or phenotypes, such as "BRCA1"                              \\
Substance                  & Chemical or biological substances involved in medicine, such as insulin and glucose                 \\
Qualifier                  & Restrictive vocabulary that modifies other entities, such as "chronic" and "acute"                  \\
Biology                    & The types of organisms involved, such as "Escherichia coli" and "influenza virus"                   \\
Specimen                   & Collect biological samples for testing, such as "blood" and "urine"                                 \\
BiologicalProcess          & Processes involving life activities, such as "cell apoptosis" and "inflammatory response"           \\
PhysicalEnergy             & Physical energy forms related to medicine, such as "X-rays" and "ultrasound"                        \\
Event                      & Specific events that occur during the medical process, such as' postoperative complications'        \\
PhysicalEntity             & Perceived and measurable physical objects, such as "blood pressure monitors" and "syringes"         \\
ObservationTarget          & Objects being observed or monitored, such as "blood glucose levels" and "temperature changes"  \\
\bottomrule 
\end{tabular}
\caption{Entity type constraints in the medical field}
\label{tab:entity_types}
\end{table}

\section{Medical Knowledge Base Extraction}
\label{appendix:med_knowledge_graph_extraction}

\textbf{Data Processing Strategy.} In this study, a hybrid semantic structural segmentation strategy is used to process the MedQA textbook data. After converting the data to Markdown format, it was split into fine-grained chunks based on chapter logic and a length limit of 1000 characters to ensure semantic integrity and preserve chapter path information. For medical website contents, we used directly structured paragraphs as independent chunks to avoid semantic fragmentation and improve retrieval efficiency.

\renewcommand\arraystretch{1.0}
\begin{table}[htbp]
\centering
\small
\setlength\aboverulesep{0pt}\setlength\belowrulesep{0pt}
\begin{tabular}{p{0.97\textwidth}}
\toprule
\prompt{\textbf{Prompt for Entity Extraction}: 

You are a highly efficient medical entity extraction system. Your task is to extract all possible entities from a given piece of medical text and generate structured data according to a unified specification. We will provide you with a piece of medical text, and based on the provided "medical type list," you need to deeply analyze the content of the text to identify and extract all relevant entities. Each entity should include fields such as name, type, domain ontology, explanation, standard name, and synonyms.

\#\#\# Output format requirements:

Please respond in the format of a JSON list, where each entity is represented as a dictionary in the following structure:

\{
    "name": "The complete term or proper noun as it appears in the original text",
    "Type": "The type of the entity selected from the type list",
    "Domain ontology": "The hierarchical classification chain of the medical domain to which the entity belongs",
    "Explain": "A descriptive explanation of the entity based on your knowledge",
    "Sandard Name": "The standard name of the entity (if different)",
    "Synonym": ["All known synonyms of the entity"]
\}

\#\#\# Medical Entity Types:

<EntityTypes>...</EntityTypes>

\#\#\# Examples:

<Examples>...</Examples>

\#\#\# Instructions for Use:

- Input a piece of medical text.

- The output should be in JSON list format, with each entry corresponding to one entity.

- Supports the extraction of multiple main entities simultaneously.

- All entities must be derived from the provided text content.

- The extracted entities can be used for subsequent applications such as knowledge graph construction, semantic search, etc.

\#\#\# Input: <Chunk>...</Chunk>
}
\midrule

\prompt{\textbf{Prompt for Relation Extraction}: 

You are an efficient medical relationship extraction system. Your task is to analyze the input text based on a given list of medical entities and extract all possible relationships that exist between these entities. The output should be in the form of a list of quadruples.

\#\#\# Each quadruple should include:

The subject entity.
A predicate representing the relationship between the two entities.
The object entity.
Specific constraints (e.g., time, location, etc.).
If no relationships are found, there is no need to list any output.

\#\#\# Output format requirements:

Respond using a JSON list format, where the entire response is a list, and each quadruple is also a list, with the format as follows:

[
  ["Subject Entity", "Predicate", "Object Entity", "Specific Constraint"],
  ...
]

\#\#\# Extraction Requirements:

1. Subject and Object Entities: Must only include entity names from the given entity list. Descriptive modifiers can be added to make the description more precise (e.g., "Insulin for intravenous injection").

2. Predicate:

- Should be meaningful terms or phrases that clearly express the relationship between the entities.

- Must have directionality, favoring concise expressions (e.g., "treats > is used to treat").

- Avoid using prepositions such as "is," "by," "in," "and," or other terms without actual semantic meaning.

- Single-word predicates should be avoided.

3. Specific Constraints:

- These are specific conditions under which the relationship between entities holds, such as time, dosage, frequency, route of administration, etc.

- Multiple constraints should be separated by commas.

- If no constraints exist, use an empty string "".

4. Source Restriction: Extracted quadruples must strictly originate from the content of the original text and not include inferred or extended relationships.

5. Formatting: Each quadruple must contain exactly four elements, separated by a comma ,.

6. Deduplication: Avoid duplicate or highly similar quadruples in the output.

7. Medical Terminology Consistency: Ensure that terminology aligns with medical standards, such as "oral administration," "intravenous infusion," or "subcutaneous injection."

<Examples>...</Examples>

\#\#\# Notice:

- All entity names must exactly match those provided in the entity list, with qualifiers added if necessary.

- Relationships must be derived directly from the text and not inferred based on common sense reasoning.

- If an entity appears multiple times in the text but the relationship is the same, retain only one instance.

- Specific details such as time, dosage, frequency, and route should be included as "specific constraints" in the fourth item.

\#\#\# Input: <Entities>...</Entities> <Chunk>...</Chunk>
}
\bottomrule
\end{tabular}
\caption{Prompt for entity and relationship extraction.}
\label{tab:med_prompt_1}
\end{table}
\textbf{Knowledge Extraction Pipeline.} Building upon the Qwen2.5-72B-Instruct model, we developed a comprehensive knowledge extraction pipeline designed to extract medical knowledge across multiple levels, including knowledge units, atomic queries, entities, and relations. This pipeline transforms unstructured text into a unified, structured knowledge base. The prompt for entity-relation extraction, as shown in Table \ref{tab:med_prompt_1}, first extracts entities based on the constraints of the specified entity types, and then extracts the relationships between these entities. For knowledge points extraction, as shown in Table \ref{tab:med_prompt_2}, knowledge points are extracted according to the prompted medical knowledge types and entity types. Each knowledge point includes core entities, related questions, and other information, thereby building a knowledge graph that comprises entities, knowledge points, related questions, and chunks.
\renewcommand\arraystretch{1.0}
\begin{table}[t!]
\centering
\small
\setlength\aboverulesep{0pt}\setlength\belowrulesep{0pt}
\begin{tabular}{p{0.97\textwidth}}
\toprule
\prompt{\textbf{Prompt for Knowledge Point Extraction}:

You are a Medical Document Analysis and Knowledge Point Extraction Assistant. Please extract knowledge points from the input document fragment (chunk) that provide a substantive description and introduction of the document's theme-related background knowledge, involved individuals, time, and location, or discuss, elaborate, or analyze relevant information in the text.

Ensure that the extracted knowledge points are strictly related to the document's theme and represent core content directly introduced, analyzed, or reasoned by the document's author in the fragment. If the knowledge point's subject cannot be clearly identified but the core entities of the document are provided in the input, you may refer to those entities.

\#\#\# Extraction Requirements

1. Identify the core discussion object in the fragment and extract content that contributes creatively to the theme of the document.

2. If the fragment contains a significant amount of referenced or mentioned content, only extract key knowledge points that involve further analysis, expansion, reasoning, or application related to those references. Do not include auxiliary information or background content.

3. The name of the knowledge point must be described in concise and precise language, fully reflecting the core theme of the knowledge point.

4. Each knowledge point should be a complete, coherent narrative or discussion. Knowledge points within a single chunk may include multiple items:

- When discussing the same theme or subject from multiple perspectives in parallel, split them into separate knowledge points.

- If further splitting compromises the causality, reasoning, or completeness of the discussion, treat the entire fragment as a single knowledge point instead of further fragmentation.

- Ensure the content description of the knowledge point is structured and avoids redundant or repetitive explanations.

\#\#\# Types of Medical Knowledge:

- Declarative Knowledge: Explanations of concepts, terms, theorems, clauses, etc., in the medical domain, answering "What is" questions (e.g., disease definitions, surgical explanations, mechanisms of drug action, examination items).

- Case-Based Knowledge: Conveying and interpreting abstract medical theories through specific clinical contexts, such as a patient diagnosed with diabetes and receiving insulin therapy.

- Factual Knowledge: Description of the state, attributes, characteristics, or background information of medical phenomena (e.g., diagnostic data, health status statistics, drug usage data).

- Procedural Knowledge: Instructions on how to perform specific medical tasks or operations (e.g., examination procedures, testing methods, surgical steps, rehabilitation training).

- Inferential Knowledge: Logic-based analysis, induction, deduction, or reasoning to arrive at medical judgments (e.g., supporting diagnostic conclusions or treatment plans).

\#\#\# Output Format:

\{
  "Knowledge Point 1 Name": \{
    "Content": "A description or summarized text fragment extracted from the chunk that unambiguously explains the knowledge point content and subject, including key information.",
    "Knowledge Type": "Declarative Knowledge/Case-Based Knowledge/Factual Knowledge/Procedural Knowledge/Inferential Knowledge",
    "Structured Content": "Structured content associated with the knowledge point, such as charts, formulas, code, rules, first-order logic representation, equation systems, data structure definitions, etc.",
    "Domain Ontology": "A hierarchical classification system of the knowledge point's medical or professional domain, consisting of broader-to-narrower category terms. The current entity must not be directly placed into the domain ontology.", "Core Entities": "Keywords representing the theme or characteristics of the knowledge unit. Each term must be semantically complete and reflect the knowledge unit's uniqueness in searches. Each core entity's valid types must correspond to the 'type list' provided.", "Related Questions": "A set of questions, fully answerable based on the knowledge point's subject and content as described in the original text.",
    "Extended Knowledge Points": "A list of additional 'most relevant' knowledge point names representing extensions of the current knowledge point. These are closely tied to the current knowledge point, follow it in a logical sequence, and are not explicitly mentioned in the original text."
  \},...
\}

\#\#\# Medical Entity Types

<EntityTypes>...</EntityTypes>

\#\#\# Input: <Chunk>...</Chunk>
}
\bottomrule
\end{tabular}
\caption{Prompt for Knowledge Point Extraction.}
\label{tab:med_prompt_2}
\end{table}

\end{document}